\documentclass[conference]{IEEEtran}
\IEEEoverridecommandlockouts

\usepackage[
backend=biber,
bibstyle=apa,
citestyle=authoryear-comp,
url=false,
isbn=false,
uniquename=false,
uniquelist=false,
mincitenames=1,
maxcitenames=2,
alldates=year,
clearlang=true,
]{biblatex}

\DeclareFieldFormat[article,unpublished,misc,preprint]{title}{``#1''}

\let\oldcite\cite
\renewcommand{\cite}[1]{{(\oldcite{#1})}}

\AtEveryBibitem{
    \clearfield{urlyear}
    \clearfield{urlmonth}
    \clearfield{urlday}
    \clearlist{language}
    \clearfield{origlanguage}
    \clearfield{url}
    \clearfield{note}
    \clearlist{location}
    \clearfield{month}
    \clearname{editor}
    \clearlist{publisher}
    \clearfield{address}
}

\DeclareFieldFormat{doi}{%
    \textsc{doi}\addcolon\space   % <--- doi instead of DOI
    \ifhyperref
        {\href{https://doi.org/\detokenize{#1}}{\texttt{\detokenize{#1}}}}
        \space
}

\usepackage{siunitx}
\usepackage{hyperref}
\usepackage[font=small,labelfont=it]{caption}
\usepackage{graphicx}
\usepackage{subcaption}
\usepackage[dvipsnames, usenames]{xcolor}
\usepackage{booktabs}

\newcommand{\mathcolorbox}[2]{\colorbox{#1}{$\displaystyle #2$}}

\let\oldsubsection\subsection
\newcommand{\subsecno}[1]{\oldsubsection*{#1}\addcontentsline{toc}{subsection}{#1}}
\renewcommand{\subsection}[1]{\subsecno{#1}}

\usepackage{ragged2e}
\usepackage{placeins}
\usepackage{cancel}

\usepackage{cprotect}
\usepackage{amsmath,amsfonts,amssymb, bm, bbm}
\usepackage{xfrac}

\newcommand{\LCC}{*_\mathrm{LCC}}
\newcommand{\had}{\circ}

\let\oldvec\vec
\renewcommand{\vec}[1]{\oldvec{#1}}
\renewcommand{\tilde}[1]{#1}
\newcommand{\hv}[1]{\mathbf{\bm{#1}}}
\newcommand{\mat}[1]{\mathbf{#1}}
\newcommand{\T}{^{\intercal}}
\newcommand{\de}{\mathrm{d}}
\newcommand{\appropto}{\mathrel{\vcenter{
  \offinterlineskip\halign{\hfil$##$\cr
    \propto\cr\noalign{\kern2pt}\sim\cr\noalign{\kern-2pt}}}}}
\newcommand{\expect}[1]{\langle #1 \rangle}

\newcommand{\wma}{\omega_\mathrm{MA}}
\newcommand{\probP}{\mathbb{P}}

\usepackage{lipsum}
\begin{document}

\title{
High-resolution spatial memory requires grid-cell-like neural codes

\thanks{
\raggedright{
\hspace*{-1.33em}\textsuperscript{1} Bio-Inspired Circuits and Systems (BICS) Lab, Zernike Institute for Advanced Materials, University of Groningen, Netherlands.\\
\textsuperscript{2} Groningen Cognitive Systems and Materials Center (CogniGron), University of Groningen, Netherlands.\\
\textsuperscript{3} Micro- and Nanoelectronic Systems (MNES), Technische Universit\"at Ilmenau, Germany.\\

\textsuperscript{4} Redwood Center for Theoretical Neuroscience, University of California, Berkeley, USA.\\
\textsuperscript{5} Energy Materials and Devices, Department of Materials Science, Kiel University, Germany.\\
Corresponding author: m.cotteret@rug.nl
}}

}

\author{\IEEEauthorblockA{\textbf{Madison Cotteret}\textsuperscript{1,2,3}, 
\textbf{Christopher J. Kymn}\textsuperscript{4}, 
\textbf{Hugh Greatorex}\textsuperscript{1,2}, \\
\textbf{Martin Ziegler}\textsuperscript{5},  
\textbf{Elisabetta Chicca}\textsuperscript{1,2},
\textbf{Friedrich T. Sommer}\textsuperscript{4}}
}

\maketitle

% I added this to get page numbers with the current document class (remove when class is changed)
\thispagestyle{plain}
\pagestyle{plain}

%%%%%%%%%%%%%%%% Text of abstract goes here %%%%%%%%%%%%%%%%%%%%%%%%%%

% \subsection*{\center \bf Abstract}

\begin{abstract}
Continuous attractor networks (CANs) are widely used to model how the brain temporarily retains continuous behavioural variables via persistent recurrent activity, such as an animal's position in an environment.
However, this memory mechanism is very sensitive to even small imperfections, such as noise or heterogeneity, which are both common in biological systems.
Previous work has shown that discretising the continuum into a finite set of discrete attractor states provides robustness to these imperfections, but necessarily reduces the resolution of the represented variable, creating a dilemma between stability and resolution.
We show that this stability-resolution dilemma is most severe for CANs using unimodal bump-like codes, as in traditional models.
To overcome this, we investigate sparse binary distributed codes based on random feature embeddings, in which neurons have spatially-periodic receptive fields.
We demonstrate theoretically and with simulations that such grid-cell-like codes enable CANs to achieve both high stability and high resolution simultaneously.
The model extends to embedding arbitrary nonlinear manifolds into a CAN, such as spheres or tori, and generalises linear path integration to integration along freely-programmable on-manifold vector fields.
Together, this work provides a theory of how the brain could robustly represent continuous variables with high resolution and perform flexible computations over task-relevant manifolds.

\vspace{1em}

\end{abstract}

\begin{refsection}

\section{Introduction}

Persistent neural activity has been proposed as a mechanism for temporarily storing behavioural variables during the execution of tasks \cite{fuster_unit_1973}.
Since the time duration demanded of such working memories typically exceeds the time constants of individual neurons, state persistence is believed to be an emergent network effect. Attractor networks are theoretical network models describing this behaviour, in which persistent activity states -- or fixed-point attractors -- are maintained by recurrent synaptic feedback \cite{little_existence_1974}. 
The dynamics in these networks can be imagined as a gradual descent of an energy function, such that energy minima correspond to stable attractor states \cite{hopfield_neural_1982, amit_modeling_1989}. Either by construction or by learning through synaptic plasticity, the energy landscape can then be tailored to support particular behaviours.

Behavioural variables are often continuous, such as head direction or one's position in space.
Accordingly, in a continuous attractor network (CAN), translational symmetries in the synaptic weights can be introduced to arrange fixed-point attractors along a low-dimensional continuum in the network's state space, such as a line.
The first generation of CANs arranged neurons in a ring, in which attractor states comprised localised \textit{unimodal} activity bumps at any angle on the ring \cite{ben-yishai_theory_1995, amari_dynamics_1977, zhang_representation_1996, skaggs_model_1995, samsonovich_path_1997}.
We call such bump-like neural representations of a behavioural variable unimodal because each neuron is selective for a single location only, like an idealized hippocampal place cell, which responds maximally near a single unique place in an animal's environment \cite{moser_spatial_2017}.

CANs have become a popular primitive for modelling biological working memory, such as for understanding navigation in the insect central complex~\cite{green_neural_2017, noorman_maintaining_2024} or in the hippocampal formation of vertebrates~\cite{chaudhuri_intrinsic_2019, gardner_toroidal_2022, burak_accurate_2009}.
However, nagging questions remain as to whether and how biological neural circuits could faithfully implement CANs. A major problem is the high degree of precision and reliability that CANs require of their individual components, which is at odds with the unreliability of neurons and synapses in the brain \cite{faisal_noise_2008, bartol_nanoconnectomic_2015}. Fast stochastic nonidealities (e.g. from stochastic spike timing) cause random walk dynamics, known as diffusion, which degrade the accuracy of the represented variable over time \cite{camperi_model_1998, compte_synaptic_2000, burak_fundamental_2012, wu_dynamics_2008, seeholzer_stability_2019, wang_multiple_2022}. Fixed nonidealities (quenched noise), such as fixed synaptic heterogeneity, lead to undesired deterministic movement of the neural state, known as drift \cite{zhang_representation_1996, seung_how_1996, renart_robust_2003}. Diffusion and drift undermine the capability of a CAN to hold or memorise a particular value, thus defeating its utility as working memory.

Introducing ``roughness'' to the energy landscape has been proposed as a solution to the diffusion problem, by creating a finite set of discrete attractor states that tile the idealised continuum. This mitigates diffusion by ``pinning'' the network to one of these states, but necessarily reduces the resolution at which a behavioural variable can be represented \cite{kilpatrick_wandering_2013, brody_basic_2003, chaudhuri_computational_2016}.
Narrowing the spacing between the discrete attractor states increases the resolution but also weakens stability, ultimately allowing diffusion between the discrete states to return \cite{kilpatrick_wandering_2013}.
This dilemma between resolution and stability in CANs with biologically-realistic nonidealities has cast doubt upon their adequacy to describe how brains represent and hold continuous variables over long timescales \cite{burak_fundamental_2012, khona_attractor_2022}.

We introduce a solution to this dilemma, which relies on using \textit{multimodal} rather than unimodal representations of continuous variables, such as in the entorhinal cortex, where grid-cell neurons are active at multiple positions in a 2D environment \cite{moser_spatial_2017, fiete_what_2008}.
Leveraging results from vector-symbolic architectures (VSAs) \cite{kleyko_survey_2022} and sketching techniques \cite{woodruff_sketching_2014}, we study CANs where the continuous variable is represented by a high-dimensional sparse binary vector using randomised multimodal codes.
We show that the noisy kernel approximations involved in these randomized coding schemes automatically introduce energy roughness, and further that they enable a CAN with traditional Hebbian plasticity to memorise continuous variables with high resolution and high robustness at the same time.
Our model suggests that multimodal entorhinal grid cells are more suitable than unimodal hippocampal place cells for supporting working memories of continuous quantities.
The proposed model also generalises to embedding arbitrary nonlinear CAN manifolds with freely-programmable integration capabilities, thus potentially explaining how general manifold-based computation could be realised in the brain~\cite{langdon_unifying_2023}.

\section{Results}

\begin{figure*}
\centering
\includegraphics[width=7in]{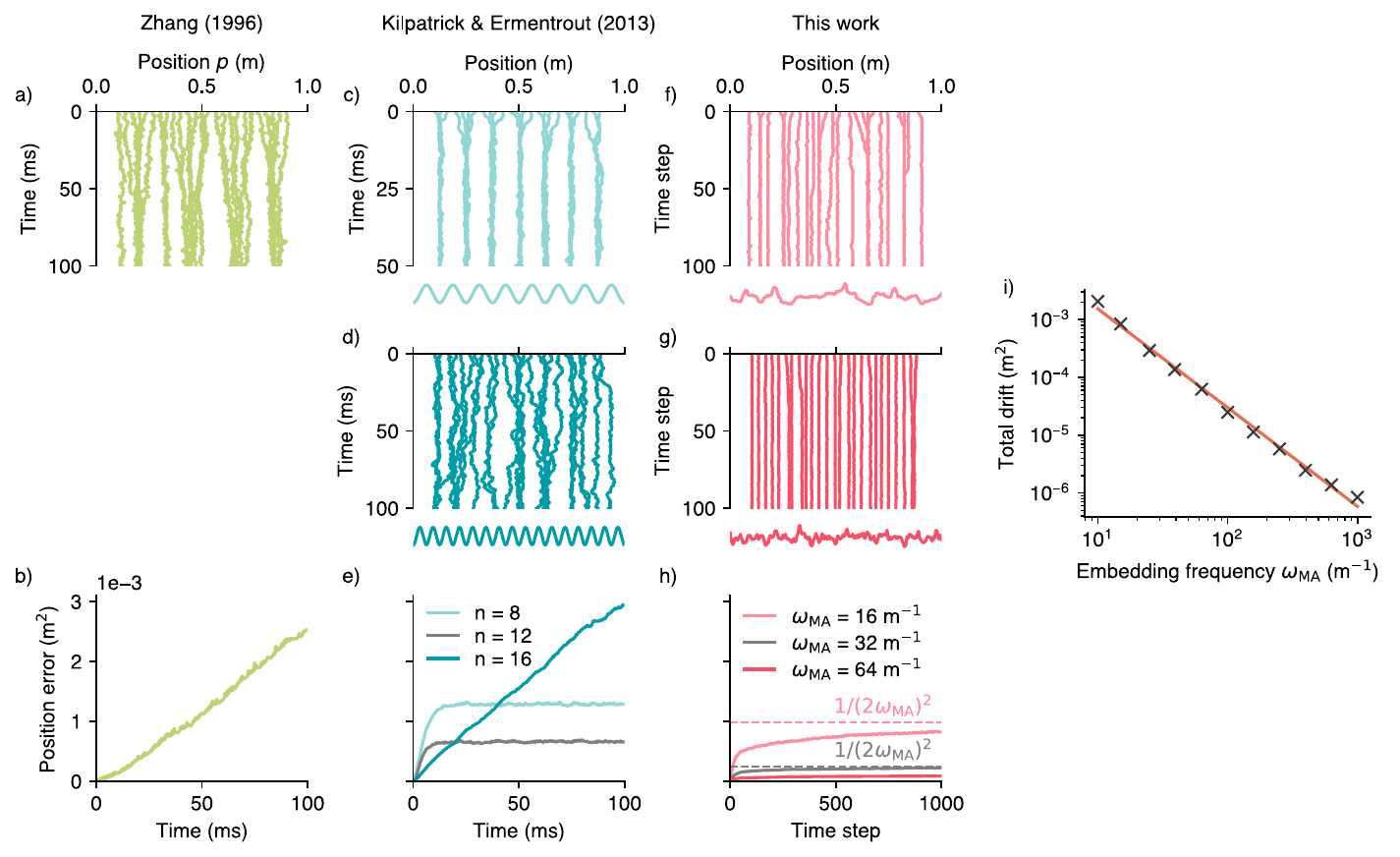}
\caption{Three line attractor models subject to equal magnitudes of nonidealities, such that \qty{1}{\ms} of noise in the first two models is approximately equal to 1 time step's worth in ours. \textbf{a)} When initiated from multiple positions, fixed synaptic nonidealities cause the model of \textcite{zhang_representation_1996} to converge on a subset of preferred states. \textbf{b)} The position error variance is linear over time, indicative of an unconstrained random walk. \textbf{c)} Introducing low-frequency periodic synaptic heterogeneity (shown below) pins the network to one of $n=8$ discrete states, mitigating diffusion between states \cite{kilpatrick_wandering_2013}.
\textbf{d)} For higher frequencies ($n = 16$) diffusion returns, and \textbf{e)} the error variance becomes linear in time again.
\textbf{f)} In our model, for a low embedding frequency ($\wma = \qty{16}{m^{-1}}$) there are large initial drifts to a subset of states, determined by the heterogeneous energy landscape (shown below).
\textbf{g)} Increasing the embedding frequency ($\wma =  \qty{64}{m^{-1}}$) minimises these drifts, but crucially does not lead to a return of diffusive random walk behaviour.
\textbf{h)} The position error variance over time for different embedding frequencies, with the length scale of the energy landscape's heterogeneity also shown. 
\textbf{i)} The long-time RMS asymptotic drift scales approximately with $\wma^{-0.85}$.
}
\label{fig:drift_multi}
\end{figure*}

\begin{figure*}
\centering
\includegraphics[width=\linewidth]{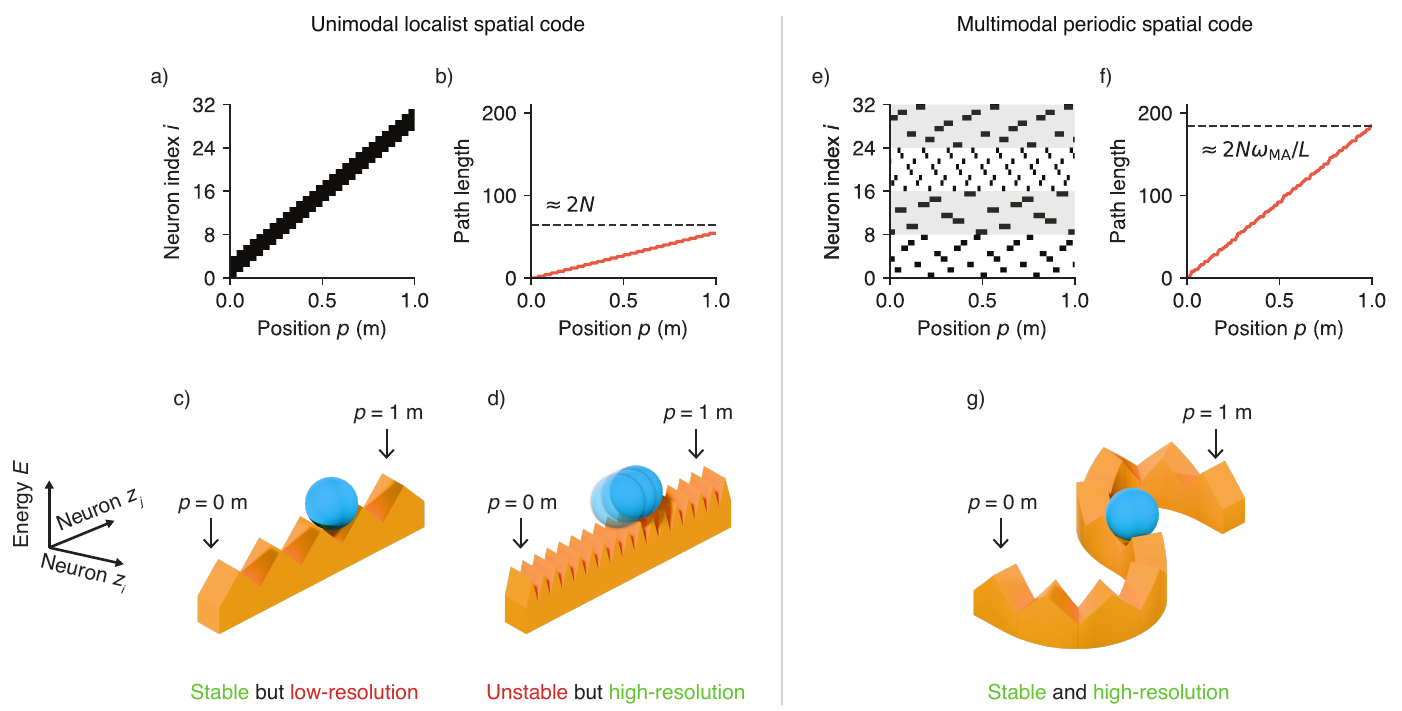}
\caption{Multimodal spatial codes embed behavioural variables into a longer path in the neural state space, enabling high-resolution and stable quasi-continuous attractor dynamics. \textbf{a)} With a unimodal spatial code, each neuron represents one position only,
\textbf{b)} limiting the $L^1$-norm path length to at most $2N$.
\textbf{c)} Tiling the line attractor with few discrete attractor states ensures stability of the network state (represented by the sphere), but limits resolution.
\textbf{d)} Increasing the number of discrete states, without increasing the neural path length, leads to stochastic transitions between states.
\textbf{e)} With a multimodal spatial code, the position representations $\hv{x}(p)$ meander cyclically through the neural state space, such that \textbf{f)} the neural path length is then dependent upon the embedding frequencies $\omega_i$.
\textbf{g)} The line attractor can then support many more discrete attractor states, without compromising their stability.
}
\label{fig:concept_fig}
\end{figure*}

To understand the principles underlying memorisation and manipulation of continuous variables in the brain, we study abstract neural network models.
Specifically, we consider an associative network of $N$ all-to-all connected binary neurons, similar to \textcite{hopfield_neural_1982}, and first describe how to embed a finite range of a 1D continuous variable $p \in \mathbb{R}$ into the network's binary state space as a line attractor. 
Although $p$ may represent any continuous behaviourally-relevant variable, we will assume it represents a spatial position between 0 and 1 metres of an animal in an environment, such as a rat's position on a 1D track.
% We thus first require an embedding function to define which states in the neural space should represent each position.

\subsection{Random spatial embeddings}

We employ an embedding function $\hv{x}(p): \mathbb{R} \rightarrow \{ 0, 1\}^N$ to define which states in the neural space should represent each position.
It should not give preference to any particular position $p$ (it is general), and it should map nearby positions to nearby points in neural space (it preserves local structure). In machine learning, randomness-based vector embeddings with these properties have been proposed and studied, such as Random Fourier Features (RFF) \cite{rahimi_random_2007}.
Similar to a quantized version of RFF \cite{raginsky_locality-sensitive_2009}, but with a fixed quantization threshold and thus sparsity, we propose an embedding function in which binary neurons have spatially-periodic receptive fields
\begin{equation}
    x_i(p) = \mathbbm{1}[ \mathrm{mod}_L (  \omega_i p + \theta_i) < 1 ]
\label{eqn:x_embedding_func}
\end{equation}
where $\mathbbm{1}[\cdot]$ is the indicator function (1 if condition true, else 0),
and $\omega_i$ and $\theta_i$ are the embedding frequency and offset of that neuron respectively.
The neuron is active for $1/ \omega_i$ metres, then inactive for $(L-1)/\omega_i$ metres, periodically.
The sparsity of the embedding is then approximately $1/L$, while $\omega_i/L$ is the approximate number of positions (modes) at which a neuron is active over the range of $p$.
The frequencies $\omega_i$ are random variables sampled from a distribution $P(\omega)$, while the offsets are sampled from a separate distribution.
The embedding (\ref{eqn:x_embedding_func}) admits a translation-invariant similarity kernel, such that the expected inner product between the vector representations of two positions $p_1$ and $p_2$ is given by
\begin{equation}
\langle \hv{x}(p_1) \cdot \hv{x}(p_2) \rangle = \frac{N}{L} K(p_1 - p_2)
\label{eqn:K_definition}
\end{equation}
where $K : \mathbb{R} \rightarrow [0,1]$ is a symmetric kernel function with a maximum at $K(0) = 1$, and falls off to $1/L$ for large differences in position.
%The kernel $K$ thus translates distances in the behavioural variable to distances in their vector representations, ensuring that the neural representation of $p = \qty{0,80}{\m}$ is very similar (close to) that of $p = \qty{0,81}{\m}$, for example.
Since an active neuron needs a change in position of approximately $\omega_i^{-1}$ metres to deactivate, the kernel width is approximately $\wma^{-1}$ metres, where $\wma = \int_{-\infty}^\infty |\omega | P(\omega) \de \omega$ is the mean-absolute embedding frequency.
The Fourier transform of $P(\omega)$ determines the exact shape of $K$.
Importantly, $\wma / L $ is the average number of modes per neuron.
Due to the kernel property (\ref{eqn:K_definition}), (\ref{eqn:x_embedding_func}) can be understood as forming a binary locality-sensitive hash of $p$ \cite{frady_computing_2022, thomas_theoretical_2022, rahimi_random_2007, raginsky_locality-sensitive_2009}.

By setting constraints on the sampled $\omega_i$ and $\theta_i$ parameters, we force the representation vectors $\hv{x}(p)$ to have a sparse block structure~(Fig. \ref{fig:concept_fig}e). Specifically, the neurons are grouped into $N/L$ blocks of integer length $L$, such that in each block exactly one neuron is active for any $p$ \cite{frady_variable_2023, laiho_high-dimensional_2015}. Thus, the resulting sparsity of the code is exactly $1/L$. Neurons in a block share one frequency $\omega_i$, sampled from $P(\omega)$, but have different phase offsets $\theta_i$.
A block in this abstract code thus resembles a grid cell module in entorhinal cortex, where grid cells share one spatial frequency but have different offsets \cite{stensola_entorhinal_2012}.

\subsection{Memory network for block-sparse spatial representations}

To store the location representations $\hv{x}(p)$ as attractors within the CAN, we use Hebb's covariance rule to form a recurrent autoassociative weight matrix \cite{sejnowski_storing_1977, hopfield_neural_1982, amari_characteristics_1989, tsodyks_enhanced_1988, palm_information_1992}
\begin{equation}
\mat{A} = \int_{p = \qty{0}{m}}^{p = \qty{1}{m}} \bar{\hv{x}}(p) \bar{\hv{x}}(p)^\intercal  \de p \ \in \mathbb{R}^{N \times N}
% \mat{A} = \int_{\text{line}} \bar{\hv{x}}(p) \bar{\hv{x}}(p)^\intercal  \de p \ \in \mathbb{R}^{N \times N}
\label{eqn:W_attr_1d} 
\end{equation}
where $\bar{\hv{x}}(p) = \hv{x}(p) - \frac{1}{L}$ are the centred position representations (\ref{eqn:x_embedding_func}), and $\mat{A}$ will be used as the synaptic weights $\mat{W}$.
To retrieve the block-sparse location representations stored in the CAN, we employ a winner-take-all (WTA) neural activation function within each block \cite{gripon_sparse_2011},
\begin{equation}
    \hv{z}_{t+1} = \mathrm{bWTA} \big[ \mat{W} \hv{z}_t \big]
\label{eqn:bwta_noinput}
\end{equation}
where $\hv{z}_t \in \{0, 1 \}^N$ are the neuron activations on time step $t$, $\mat{W} \in \mathbb{R}^{N \times N}$ is the synaptic weight matrix,  and $\mathrm{bWTA}$ is a per-block WTA function.
A potential biological mechanism for implementing the $\mathrm{bWTA}$ function is fast competition through inhibitory connectivity within a block or module, like divisive normalization \cite{shim_statistical-mechanical_1991, heeger_normalization_1992}.

Setting the block size $L$ in the embedding to a large integer value yields sparse representations of position, in line with the sparse activity of biological networks.
Further, in networks with Hebbian learning, sparse patterns exploit synaptic memory capacity most efficiently, that is, they maximize the number of uncorrelated discrete attractor states a network can support \cite{amit_information_1987, amari_characteristics_1989, tsodyks_enhanced_1988, palm_information_1992}. Interestingly, the block constraint introduced by (\ref{eqn:bwta_noinput}) retains this high synaptic memory capacity
\cite{gripon_sparse_2011, knoblauch_iterative_2020}.
While direct application of such capacity results is here complicated by the fact that we are storing a quasi-continuum of correlated states, they nonetheless provide a high upper capacity bound when using sparse codes.

We impose nonidealities upon the network in two ways. First, we add fixed random noise to the synaptic weights, to imitate fixed synaptic heterogeneity. Second, we introduce a stochastic degradation of the neural state, by randomly setting on each time step 10\% of the neurons to either a 0 or 1 state, with equal probability. This relatively high noise value is tempered by attractor dynamics, which project the perturbation back onto the attractor manifold, reducing its impact by a factor of $ \sqrt{N}$ \cite{burak_fundamental_2012, wu_dynamics_2008, compte_synaptic_2000}.

\subsection{The resolution-stability dilemma in traditional CANs}

We compare the stability and resolution of three line attractor models with comparable imposed nonidealities: the neural field models of \textcite{zhang_representation_1996} and \textcite{kilpatrick_wandering_2013}, which use unimodal place-cell-like codes, and the model introduced in this work, which uses multimodal periodic codes (\ref{eqn:x_embedding_func}).
In the first model, synaptic heterogeneity causes the idealised continuum of states to uncontrollably deteriorate into a set of discrete attractor states. When the network is initialised from multiple positions along the line attractor, there are large initial drifts to this reduced set of positions (Fig.~\ref{fig:drift_multi}a). The network state is not confined to these states however, and the stochastic neural dynamics then cause the network state to perform a diffusive random walk along the line attractor (Fig.~\ref{fig:drift_multi}b), where the position variance is proportional to time.

To mitigate diffusion of the behavioural variable over time, the model of \textcite{kilpatrick_wandering_2013} introduces sinusoidal heterogeneity of frequency $n$ along the line attractor, to deliberately collapse the continuum of states to a set of $n$ equally-spaced discrete attractor states. For low $n$, there are large initial drifts as the network state self-discretises into one of the $n$ attractor states (Fig. \ref{fig:drift_multi}c). Afterwards, the network state is ``pinned'' to that particular attractor state, mitigating diffusion (Fig. \ref{fig:drift_multi}e).
The network is thus stable, but it has low resolution.

Increasing $n$ increases the number of discrete attractor states, thereby increasing the resolution. However, this necessarily also decreases the spacing between the attractor states.
The result is that, for even moderately larger $n$ values, the distance between the would-be attractor states is small enough to allow stochastic transitions between them \cite{kilpatrick_wandering_2013}. Diffusion of the behavioural variable returns, and is comparable in magnitude to the first model \mbox{(Fig. \ref{fig:drift_multi}d,e)}.
This presents a severe trade-off between the resolution of the represented behavioural variable and its stability \mbox{(Fig. \ref{fig:concept_fig}c,d)}.

\subsection{Escaping the resolution-stability dilemma}

In the absence of noise, the block-sparse memory network will approximately descend the energy $E = -\hv{z}^\intercal \mat{W} \hv{z}$ (see Methods). Although the energies $E(p)$ of states on the line attractor (\ref{eqn:W_attr_1d}) have equal expected values, the actual energy as a function of location $E(p)$ is not flat, but instead consists of a rough landscape of minima and maxima. This is because with the randomized embedding (\ref{eqn:x_embedding_func}), the inner products between location representations $\hv{x}(p_1) \cdot \hv{x}(p_2)$ are not equal to the kernel (\ref{eqn:K_definition}) exactly, but are themselves random variables.
The roughness of $E(p)$ has an autocorrelation length scale of $1/(2 \wma)$ metres, where $\wma = \int_{-\infty}^\infty |\omega | P(\omega) \de \omega$ is the mean-absolute embedding frequency; therefore, we expect the drifts to occur over the same length scale.

As in the Kilpatrick \& Ermentrout model, our model initially drifts to a stable local minimum, with an average drift distance of $1/(2 \wma)$ metres \mbox{(Fig. \ref{fig:drift_multi}f)}.
To increase the CAN's resolution, we thus increase $\wma$ to increase the number of such stable locations.
Interestingly, however, when the spatial resolution is increased in this way, the behaviour of the two models diverge qualitatively. In our model, the stable memory function persists, whereas in the Kilpatrick \& Ermentrout model diffusion returned and the memory function was compromised \mbox{(Fig. \ref{fig:drift_multi}d)}. The average position errors in the two models for different resolutions can compared in \mbox{(Fig. \ref{fig:drift_multi}e,h)}.

Why, then, is our model not subject to the same resolution-stability dilemma as the comparison models in Fig.~\ref{fig:drift_multi}? 
To answer this, we define a quantity describing embeddings of continuous variables in vector spaces: the {\it path length in representation space}, or just path length. For binary embeddings the $L^1$-norm path length is the sum of Hamming distances along the vector embedding as the continuous variable sweeps its entire range.
% it is the sum of Hamming distances or the number of neural switches that happen as the continuous variable is gradually moved through its entire range. 
The comparison models in Fig. \ref{fig:drift_multi} are bump attractor networks, in which each neuron has a unimodal receptive field over the range of the continuous variable. As $p$ sweeps through its range, each neuron becomes active at its preferred position, then inactive again. Thus, each neuron changes state at most twice and the resulting path length is about $2N$ \mbox{(Fig. \ref{fig:concept_fig}a,b)}.
Any increase in the number of discrete states along the line attractor must thus reduce the distance between them.

In contrast, for neurons with multimodal receptive fields, like in (\ref{eqn:x_embedding_func}), each neuron changes state two times for each mode, and so the path length is $2N$ times the average number of modes $\wma /L$ \mbox{(Fig. \ref{fig:concept_fig}e,f)}.
By increasing $\wma$ and thus the number of modes, we map the line attractor to a longer path in the neural state space.
The line attractor can then support more discrete attractor states while keeping the distance between them in the neural state space constant.
This avoids inadvertently reintroducing diffusion between states, and enables memory of a behavioural variable with high resolution and high stability.

By sweeping $\wma$, we find that the root-mean-square (RMS) drift is proportional to $\wma^{-0.85}$ \mbox{(Fig. \ref{fig:drift_multi}i)}. If the only change with increasing $\wma$ were to increase the attractor path length, without disturbing the dynamics elsewhere on the attractor, then we would expect the RMS drift to be proportional to $\wma^{-1}$. This discrepancy can be attributed to the increased memory loading and thus interference from storing more states within the network, as in conventional discrete attractor models \cite{hopfield_neural_1982, amit_modeling_1989}. Nonetheless, these results show that a 4096-neuron line attractor utilising periodic spatial codes can accurately represent a metre-scale position to millimetre resolution, despite considerable imposed neural and synaptic nonidealities.

\subsection{Adding switchable path integration}

\begin{figure*}[h!]
\centering
\includegraphics[width=0.95\linewidth]{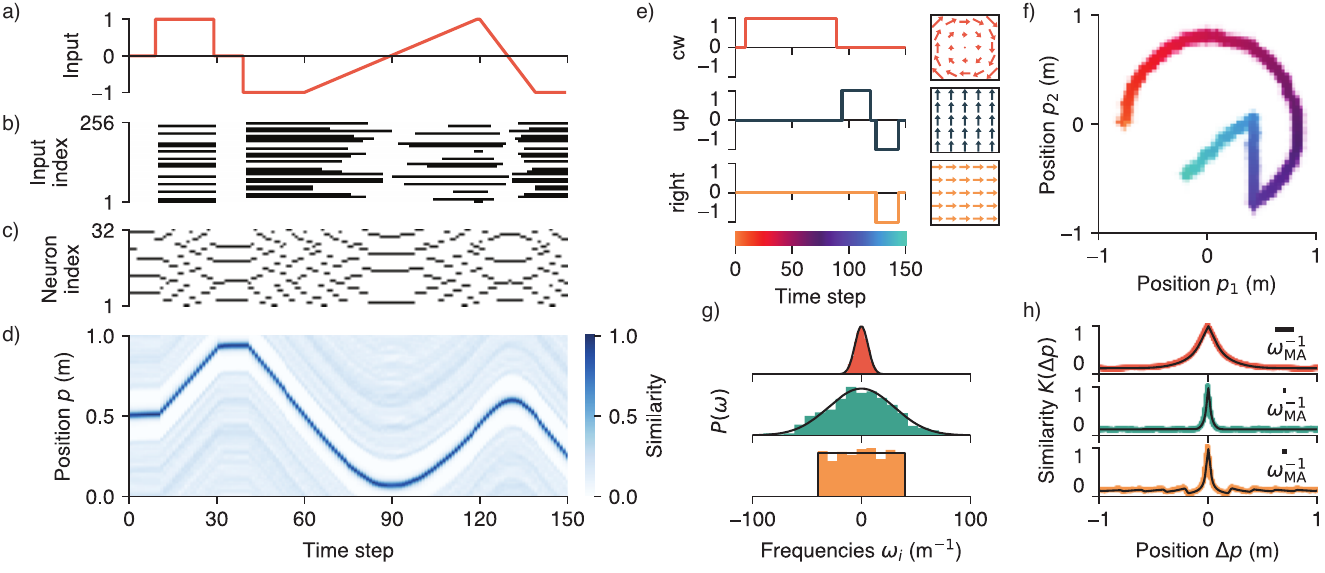}
\caption{Input integration in a 1D line and 2D plane CAN. \textbf{a)} Input to a line attractor network and \textbf{b)} the corresponding inhibitory mask applied. Only the first 256 of 4096 neurons are shown. \textbf{c)} The binary network state $\hv{z}_t$ over time, with block length $L = 8$. \textbf{d)} The similarity between $\hv{z}_t$ and the position vectors $\hv{x}(p)$. The maximally-similar state indicates the position represented by the network. \textbf{e)} Inputs to a 2D plane attractor. Each input corresponds to a different vector field and inhibitory mask. \textbf{f)} The position represented by the attractor network, with time indicated by colour. \textbf{g)} Histograms of sampled embedding frequencies and \textbf{h)} their measured and theoretical (in black) similarity kernels.}
\label{fig:integration_megafig}
\end{figure*}

Path integration is the ability to gradually update a represented location using velocity information \cite{khona_attractor_2022}.
It is a common computational motif in biological systems (e.g. in the Drosophila head direction circuit) and enables behaviours such as navigating in the dark and path planning \cite{green_neural_2017}. It is also theorised to underpin more general manifold-based computation in biological networks \cite{langdon_unifying_2023}.

To enable path integration in the CAN, heteroassociative weights $\mat{H}$ may be superimposed upon the autoassociative weight matrix (\ref{eqn:W_attr_1d}), $\mat{W} = \mat{A} + \mat{H}$.
The added term should release the network from its autoassociative dynamics, such that the current state $p$ should instead project to a shifted position state $p + \delta p$ \cite{zhang_representation_1996, burak_accurate_2009}.
The $\mat{H}$ matrix should effectively project out the difference between consecutive $\hv{x}(p)$ states, and so is formed by integrating over the outer product between the vectors $\frac{\de \hv{x}}{\de p}$ and $\bar{\hv{x}}$ over the path of $p$,
\begin{equation}
    \mat{H} =  c \int_\text{line} \frac{\mathrm{d} \hv{x}}{\de p}  \big( \bar{\hv{x}} \had  \hv{s} \big)\T \: \de p
\label{eqn:H_matrix_1d}
\end{equation}
where $c \in \mathbb{R}$ is a constant determining the maximum integration speed. The 
vector $\hv{s} \in \{ -1, 1\}^N$ contains random bipolar entries, and is \textit{bound} to $\bar{\hv{x}}$ via the Hadamard product ``$\had$'' (element-wise multiplication).
Binding $\hv{s}$ to $\bar{\hv{x}}$ ensures that, in the absence of velocity input, the rows of $\mat{H}$ have 0 expected inner product with any position state $\hv{x}$, such that no integration behaviour occurs, i.e. $\mat{W} \hv{x}(p) \approx \hv{x}(p)$. To activate path integration, external input inhibits all neurons for which $s_i = -1$. Then, these inner products become nonzero, such that the postsynaptic sum approximates $\hv{x}(p+c)$ rather than $\hv{x}(p)$, giving rise to integration dynamics. This is a ``push-pull'' path integration mechanism, wherein an equal number of neurons are biased in the positive and negative directions, resulting in no net bias in the absence of input \cite{burak_accurate_2009}.

Path integration with a 1D line attractor is shown in Fig.~\ref{fig:integration_megafig}a-d. When one subset of neurons is masked by input, the attractor moves in the positive $p$ direction, while inhibiting the opposite set of neurons triggers integration in the negative $p$ direction. If fewer neurons are masked by the input, the integration is correspondingly slower.
Importantly, this method of introducing integration behaviours by superimposing heteroassociative matrix terms (\ref{eqn:H_matrix_1d}) easily generalises to higher-dimensional attractor manifolds with multiple inputs.

\subsection{Higher-dimensional CAN manifolds}

\begin{figure*}
\centering
\vspace{-1em}
\includegraphics[width=4.5in]{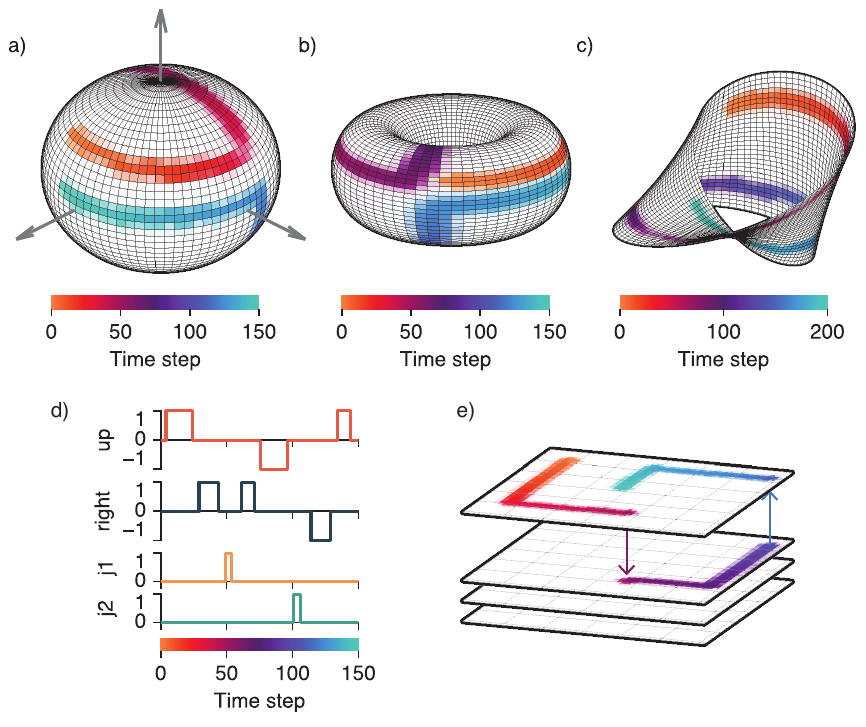}
\caption{\textbf{a-c)} Attractor networks with different embedded manifolds and binary weights integrate inputs along their respective vector fields. In particular, the sphere attractor's vector fields are rotations around each axis.
While the sphere and torus attractors have clear application for spatial representation \cite{cannon_proposed_1983, samsonovich_path_1997, mcnaughton_path_2006}, the Möbius strip highlights the generality of our approach, and that it is not reliant on geometric symmetries.
\textbf{d)} The inputs to and \textbf{e)} the position represented by an attractor network with multiple manifolds embedded simultaneously. The inputs ``up'' and ``right'' are associated with orthogonal vector fields, while the inputs ``j1'' and ``j2'' cause transitions between manifolds.
}
\label{fig:crazy_manifolds}
\end{figure*}

Storing higher-dimensional manifolds in our CAN requires only that we modify the scalar position embedding function $\hv{x}(p)$ (\ref{eqn:x_embedding_func}) to a vector embedding function $\hv{x}(\vec{p}) : \mathbb{R}^D \rightarrow \{0, 1\}^N$, and perform the weight matrix integrals over the appropriate number of dimensions. We use the arrow vector notation $\vec{p} = (p_1, p_2, \ldots )\T$ to denote low-dimensional vector quantities to be embedded, distinguishing them from the high-dimensional random vector representations $\hv{x}$. A convenient way to construct $\hv{x}(\vec{p})$ is to employ $D$ separate scalar embeddings as in the 1D case, and bind these vectors together \cite{frady_variable_2023}. The vectors $\hv{x}(\vec{p})$ then admit a $D$-dimensional translation-invariant kernel $K: \mathbb{R}^D \rightarrow [0,1]$ analogous to the 1D case (\ref{eqn:K_definition}) (\ref{sec:SM_kernel_multi}).

To implement path integration on a 2D plane attractor, the $\mat{A}$ and $\mat{H}$ matrices are constructed by integrating over two position dimensions. In the 1D line attractor case, the matrix $\mat{H}$ projected to the derivative column vector $\frac{\de \hv{x}}{\de p}$. With a higher-dimensional manifold embedded, one can freely choose to project to any linear combination of the partial derivatives $\frac{\partial \hv{x}}{\partial p_{1}}$ and $\frac{\partial \hv{x}}{\partial p_{2}}$ at every point on the manifold.
Different vector fields associated with different $\hv{s}$ vectors may then be superimposed, each triggered by a different pattern of inhibition. In \mbox{Fig.~\ref{fig:integration_megafig}e,f} we construct a plane attractor for which one input causes rotation about the origin, and two others movement in orthogonal directions. By programming arbitrary vector fields onto the manifold, the network is able to perform general nonlinear integration of multivariate inputs.

Path integration can be equally performed on manifolds with more complex topologies.
To store a ring attractor, for example, we could simply integrate the position states over a ring in 2D space, rather than the whole plane.
In Fig. \ref{fig:crazy_manifolds} we construct continuous attractor networks with sphere, torus, and Möbius strip topologies, with associated vector fields.
We stochastically binarised the weights in these networks, to highlight that they can be implemented in neural systems with low synaptic precision (Fig.~\ref{fig:crazy_manifolds}a-c).

\subsection{Multiple manifold embeddings}

Storing multiple manifolds in a single attractor network is thought to underlie the brain's ability to store representations of multiple environments \cite{wills_attractor_2005, samsonovich_path_1997}.
As with the 1D line attractors, networks storing multiple manifolds using unimodal spatial codes encounter issues of severely limited spatial resolution \cite{battaglia_attractor_1998, cerasti_spatial_2013, monasson_crosstalk_2014, battista_capacity-resolution_2020}.
However, in our model, multiple manifolds may be simultaneously embedded simply by superimposing the weight matrices responsible for each separate manifold (Fig.~\ref{fig:crazy_manifolds}e).
The separate matrices may be generated by resampling the embedding parameters $\omega_i$ and $\theta_i$ (\ref{eqn:x_embedding_func}) for each manifold.
This simple scheme has a useful property: the cross-talk noise from states on different manifolds is then no different from between distal points on the same manifold.
Additional heteroassociative terms in the weight matrix can be superimposed in order to trigger transitions between manifolds, similar to previous works embedding finite state machines into discrete attractor networks \cite{cotteret_vector_2024, cotteret_distributed_2025}.

\section{Discussion}

\subsection{High-resolution \& stable working memory}

Recent theoretical work on the \textit{Drosophila} navigation system has suggested modelling biological working memories of continuous behavioural variables with CANs composed of a small number of neurons \cite{noorman_maintaining_2024}.
In agreement with past literature, Noorman \textit{et al.} suggest that, much like digital computers, some degree of discretisation at the neuron \cite{goldman_robust_2003, koulakov_model_2002, kitano_time_2003, loewenstein_temporal_2003, camperi_model_1998} or network level \cite{kilpatrick_wandering_2013, kilpatrick_optimizing_2013, pouget_digitized_2002, brody_basic_2003, miller_analysis_2006} is required to maintain stable representations in biological memory circuits.

To better understand the resulting resolution-stability dilemma, we defined the path length of the encoded variable in representation space. For binary neurons, the path length is easily measured by the Manhattan distance. If the path length is very short, then the behavioural variable is tightly packed into a very small portion of the representation space, such that only very few well-separated stable points can exist along the path (stable but low-resolution). This is the case when using unimodal place-cell-like neural codes, like those in conventional bump attractor networks, where the path length is at most $2N$ \cite{skaggs_model_1995, zhang_representation_1996, kilpatrick_wandering_2013}. In contrast, the path length with a multimodal grid-cell-like code (\ref{eqn:x_embedding_func}) is dependent upon the spatial frequency of its receptive fields, and so may far exceed $2N$. A CAN using multimodal grid-cell-like codes may thus stably represent behavioural variables with significantly greater resolution than one with unimodal place-cell-like codes.
This result should extend to networks with more complex neuron models, e.g. continuous-valued neurons, as long as the definition of path length is suitably adjusted.

The representation path length may also provide insight into why entorhinal 2D grid cells use grid-like receptive fields at all, rather than plane-wave receptive fields, which would be another natural choice \cite{sorscher_unified_2023}. An active grid cell alters its activity for movements in all directions, while a plane-wave-coded neuron does not change its state for movements parallel to the wave front. Plane-wave codes may thus fail to adequately separate position representations in neural space, leading to a resolution-stability trade-off that is similar to, but somewhat milder than, that suffered by place-cell codes.

\subsection{Models of the hippocampal formation}

There have historically been two opposing views of the essential function of the hippocampal formation: some have emphasized its role in spatial navigation via the concept of cognitive maps~\cite{okeefe_hippocampus_1978, samsonovich_path_1997, behrens_what_2018}, while others have proposed a broader role as an associative memory~\cite{eichenbaum_role_2017, treves_computational_1994}.  There have been many attempts to unify these perspectives, suggesting a general function as a relational memory system, not limited to spatial navigation per se~\cite{whittington_tolman-eichenbaum_2020, eichenbaum_can_2014, buzsaki_memory_2013, chandra_episodic_2025}. Our model is a concrete mechanistic instantiation of this unified view. We show that an associative memory with standard Hebbian learning for memorising sparse distributed random vectors can function as a spatial memory, and, if equipped with integration capabilities, as a spatial navigation system.
% \color{red}
This points to the entorhinal cortex, with its distributed place representations, as a site for continuous attractor dynamics, as opposed to hippocampal CA3 as proposed in early modelling works \cite{samsonovich_path_1997, kali_involvement_2000, li_mechanisms_2024}.

If the hippocampal formation is indeed optimized for both navigation and memory,
then how does a generic encoding of position in entorhinal cortex contribute to this function?
% it is still an active research topic how an abstract encoding of position in entorhinal cortex contributes to this function.
Recent work has emphasized the compositional nature of computation in the hippocampal formation, in that it manipulates data structures that combine diverse behaviourally-relevant information \cite{kurth-nelson_replay_2023}.
In cognitive neuroscience, vector-symbolic architectures (VSAs) were proposed as a framework for such compositional neural computation  \cite{plate_holographic_2003, kanerva_hyperdimensional_2009, gayler_multiplicative_1998} and, as a historical aside, also anticipated the ideas used here for encoding spatial position \cite{plate_holographic_1992}.
In VSAs, symbols and positions are represented by high-dimensional random vectors, which can be combined through multiplicative ``binding'' operations to express role-filler pairs and through additive ``bundling'' operations to form sets.
These operations equip neural networks with sufficient syntax to operate over compositional data structures and perform tasks like analogical reasoning~\cite{kleyko_survey_2022}. %
Concretely, a recent model of the hippocampal formation proposes vector binding between generic spatial codes in medial entorhinal cortex (MEC) and sensory representations in lateral entorhinal cortex (LEC) to generate a cognitive map, serving as a memory index for hippocampus~\cite{kymn_binding_2024}. 
Our model takes steps towards a mechanistic neural implementation of these ideas.
Specifically, we propose how to address the resolution-stability dilemma that riddles any attempt to model spatial encoding in MEC, and we show how this functionality can be robustly implemented with imperfect and noisy components.

\subsection{Programmable neural manifolds}

There is growing evidence that low-dimensional neural manifolds are a recurring computational primitive in the brain \cite{langdon_unifying_2023, vyas_computation_2020}.
Far from being limited to spatial navigation, reports over the last decade indicate that both near-sensory and high-level cognitive behavioural variables occupy low-dimensional manifolds in the neural representation space, such as social distance \cite{nieh_geometry_2021, tavares_map_2015}, emotional states \cite{nair_approximate_2023}, object perception \cite{constantinescu_organizing_2016, behrens_what_2018} and motor commands \cite{gallego_neural_2017}. While low-rank connectivity gives rise to linear manifold representations~\cite{mastrogiuseppe_linking_2018, darshan_learning_2022, pollock_engineering_2020, mante_context-dependent_2013}, less is known about general ways to confine neural activity to task-appropriate nonlinear neural manifolds.

Notably, the approach proposed in this work generalises seamlessly to generating structure-preserving neural embeddings of arbitrary nonlinear neural manifolds, and CANs which denoise, memorise, and manipulate points on the chosen manifold.
The nonlinear similarity kernel (\ref{eqn:K_definition}) also allows us to freely ``program'' the direction in which the state should move for every input and location.
Combined, this endows CANs with a tremendous degree of flexibility as a computational primitive: if a behaviour or computation can be expressed in terms of controllable flows on low-dimensional manifolds, then it can be implemented by a CAN.
For example, to manipulate 3D rotational velocity signals, such as in the vestibular system \cite{seung_how_1996, cannon_proposed_1983}, a sphere attractor with rotational vector fields as in Fig.~\ref{fig:crazy_manifolds}a would be suitable, while a torus would enable the boundary-free manipulation of 2D positions \cite{mcnaughton_path_2006, gardner_toroidal_2022, chaudhuri_intrinsic_2019}. Even going beyond single manifolds, one can construct CANs that encode multiple manifolds (Fig.~\ref{fig:crazy_manifolds}e). Such a network could then dynamically switch to the manifold best suited to manipulate a given input, as has been proposed for solving tasks with compositional structure \cite{driscoll_flexible_2024, yang_task_2019, duncker_organizing_2020}.

\color{black}

\section{Methods}

\subsection{Scalar position encoding}

The position representation vectors $\hv{x}(p)$ are generated according to (\ref{eqn:x_embedding_func}). Instead of indexing the neurons by $i = 0 \ldots N-1$, we can instead index them by their block index $m = \lfloor i /L \rfloor$ and within-block index $n = \mathrm{mod}_L \: i$, such that $i = n + Lm$. For every block $m$, a single frequency value $\omega_m$ is independently sampled from $P(\omega)$, and all neurons in the block have this frequency value. The random offsets $\theta_{m,n}$ are then set such that exactly one neuron in each block is active for any position $p$. They are given by $\theta_{m,n} = \pi^{(m)}_{n} + \theta_{m}$ where $\vec{\pi}^{(m)}$ is a random permutation of $0 \ldots L-1$ and $\theta_{m} \overset{d}{=} \mathcal{U}_{[0,1]}$ is a uniformly distributed random offset. Both the permutation vector and offset are independently generated for each block. The position states for 4 blocks of neurons with block length $L= 8$ are shown in Fig. \ref{fig:concept_fig}e.

Deriving the similarity kernels requires calculating the expectation $\expect{x_i (p) x_i(p + \Delta p)}_{\omega_i, \theta_i}$ at any index $i$, yielding
\begin{equation}
K(\Delta p) = \frac{1}{L} + \frac{2}{L} \sum_{n = 1}^{\infty}  \mathrm{sinc}^2 \Big( \frac{n}{L} \Big)  \hat{P}\Big(\frac{2 \pi  n}{L} \Delta p \Big)
\end{equation}
where $\hat{P}(u) = \int P(\omega)\exp(-i \omega u) \de \omega$ is the Fourier transform of $P(\omega)$ (\ref{sec:SM_kernels}).
This directly connects our embedding function's (\ref{eqn:x_embedding_func}) kernel and those of RFF phasor embeddings \cite{rahimi_random_2007}, where the kernel is simply $\hat{P}(\Delta p)$, and their quantised variants \cite{raginsky_locality-sensitive_2009, li_signrff_2022}.
The measured and theoretical kernels for three different frequency distributions $P(\omega)$ are shown in Fig. \ref{fig:integration_megafig}g,h. For large $\Delta p$ the kernels approach $1/L$. We thus use the notation $\bar{K} = K - 1/L$ to denote the unbiased kernel, which tends to $0$ for large $\Delta p$.

% To calculate statistics relating to the heterogeneity of the energy landscape,
We consider in detail the dynamics of inner product $X(p) := \hv{x}(p) \cdot \hv{y}_0 $ where $\hv{y}_0$ is a fixed independent sparse block vector. This can be approximated by the stochastic differential equation
\begin{equation}
    \de X = - \wma \frac{L}{L-1} \Bigg[ \frac{N}{L^2} - X \Bigg] \de p + \sqrt{\frac{2N}{L^2} \wma} \de W
\label{eqn:OU_approx}
\end{equation}
where $\de W$ is an increment of a Wiener process, with $\expect{\de W} = 0$ and $\expect{\de W^2} = \de p$ (\ref{sec:SM_stochproc}). 
Using (\ref{eqn:OU_approx}), we can derive statistics of the heterogeneity of the energy landscape $E(p)$, most importantly the autocorrelation length scale (\ref{sec:SM_stochproc}).

\subsection{Network architecture}

All the networks introduced in this paper are single layer RNNs. For experiments without integration, the weight matrix consists only of autoassociative terms. The weight between neurons in the same block is set to 0, equivalent to the stipulation of no self-connections in Hopfield networks \cite{hopfield_neural_1982, amit_modeling_1989}. For the 1D line attractor $\mat{A}$ is constructed as defined by (\ref{eqn:W_attr_1d}), calculated by a Riemann sum with a sufficiently-small step size. This matrix ensures that the active position state is maximally reprojected out within the postsynaptic sum, since
\begin{equation}
\begin{split}
\hv{x}(q) \cdot \big( \mat{W} \hv{x}(p) \big) & = \int_\text{line} \big[ \hv{x}(q) \cdot \bar{\hv{x}}(p^\prime) \big]  \big[ \bar{\hv{x}}(p^\prime) \cdot \hv{x}(p) \big] \de p^\prime \\
& \approx \int_{-\infty}^\infty  \bar{K}(q - p^\prime) \bar{K}(p -p^\prime) \de p^\prime
\end{split}
\end{equation}
for a comparison position $q$. We have assumed that $p$ is far from the edges of the line attractor, and that the inner products between position states are well-approximated by their expected values. By the Cauchy-Schwarz inequality, this integral is maximised for $p = q$, i.e. the postsynaptic sum is most similar to the current position.

The block-WTA activation function ensures that only the neuron with the greatest input in each block becomes active. For example, for block length $L=2$, $\mathrm{bWTA}[(3,4,-5,-2)] = (0, 1, 0, 1)$.
Since the postsynaptic sum is most similar to the current position state, the $\mathrm{bWTA}$ operation approximately cleans up $\mat{W}\hv{x}(p)$ to $\hv{x}(p)$, realising the desired attractor dynamics.

\subsection{Synaptic and neural nonidealities}

We introduced fixed synaptic nonidealities to a weight matrix $\mat{W}$ by superimposing Gaussian noise of equivalent magnitude to the ideal weight values, via
\begin{equation}
    w_{ij}^\text{nonideal} = w_{ij}^\text{ideal} + \chi_{ij} \cdot w^\text{ideal}_\mathrm{RMS}
\end{equation}
where $\chi_{ij}$ are samples from a standard Gaussian distribution, and are scaled by $w^\text{ideal}_\mathrm{RMS}$, the RMS of the ideal weight values.
Fast stochastic neural nonidealities were introduced by imposing a bit error rate $b = 10\%$, such that on every time step, each neuron's state was set to $0$ or $1$ with equal probability $\frac{1}{2} b$, or left unaltered with probability $1 - b = 90\%$.

In Fig. \ref{fig:crazy_manifolds}, the weights were stochastically binarised via
\begin{equation}
w_{ij}^\text{bin} = \begin{cases}
    1 \: \: \text{w.p.} \: \: ( 1 + \exp [ - \alpha ( w_{ij} - \langle w \rangle ) / \sigma_w ] )^{-1} \\ 0 \: \: \text{otherwise}
\end{cases}
\end{equation}
where $\expect{w}$ and $\sigma_w$ are the mean and standard deviation of the ideal weight values respectively, and $\alpha$ is the steepness of the sigmoid, which we set to $\alpha = 5$.

\subsection{Energy function}
\label{sec:energy}

We consider that our network is following a discrete-time approximation with Euler time step $\tau$ of the continuous-time dynamical system
\begin{equation}
    \tau \frac{\de \hv{z}}{\de t} = -\hv{z}(t) + \mathrm{bWTA} [ \mat{A} \hv{z}(t)]
\end{equation}
where the weights contain only the autoassociative component $\mat{A}$, in which $\hv{z}$ decays to $\mathrm{bWTA}(\mat{A}\hv{z})$ in time $\tau$, rather than jumping directly to it. We can then define a scaled energy
\begin{equation}
E(\hv{z}) = -\frac{ L^2 \wma}{N}\hv{z}\T \mat{A} \hv{z}
\end{equation}
which, due to $\mat{A}$ being symmetric has time derivative
\begin{equation}
\Big( \frac{\tau N}{L^2 \wma} \Big) \frac{\de E}{\de t} = \big( \mat{A} \hv{z} \big) \cdot \big( \hv{z} - \mathrm{bWTA}(\mat{A} \hv{z})  \big) \leq 0
\end{equation}
such that $E$ is never-increasing, and is thus a Lyapunov function of our dynamics \cite{cohen_absolute_1983, shim_statistical-mechanical_1991}.
When the heteroassociative terms (or also random synaptic heterogeneity) are added to $\mat{W}$, we break the symmetry assumption, and so the descent on $E$ no longer strictly holds.
It is nonetheless useful for describing the network dynamics when there is no input, assuming that the bound heteroassociative terms contribute only a weak noise to the postsynaptic sum.
Using (\ref{eqn:OU_approx}) we can derive expressions for the statistics of $E(p)$. Most importantly, the autocovariance can be approximated by
\begin{equation}
    \mathrm{cov} \big[ E (p) , E (p + \Delta p) \big] \approx \big( 4N \kappa_1^2 + \wma \big) e^{- 2 \wma \Delta p}
\label{eqn:covariance_methods}
\end{equation}
where $\kappa_1 = \int_{-\infty}^\infty \bar{K}(u/ \wma) \de u$ is independent of $\wma$ (\ref{sec:SM_stochproc}). Critically, (\ref{eqn:covariance_methods}) implies that the energy landscape's heterogeneity has an autocorrelation length scale of $1/(2 \wma)$ metres, which we interpret as the effective spacing between discrete attractor states.

% \subsection{Position readout}

% The position represented by the network $\hat{p}$ was taken to be the position state $\hv{x}(p)$ for which the similarity was maximised, i.e.
% %
% \begin{equation}
% \hat{p}_t = \mathrm{argmax}_{p} \: \hv{x}(p) \cdot \hv{z}_t
% \end{equation}
% %
% where $\hv{z}_t$ is the network state on time step $t$. A valid criticism of this approach is that it does not scale favourably for increasing manifold dimension, requiring an exponential number of inner products to be evaluated. If desired, more efficient vector decoding mechanisms could be employed, however \cite{frady_resonator_2020, kymn_computing_2024, hersche_factorizers_2024, raviv_linear_2024, renner_neuromorphic_2024}. 

% To generate the figures showing the overlaps superimposed over time, the maximum overlap and time at which it occurred was measured for each position tuple. The time determined the base colour value for each position according the colour map, in HSV colour space. The saturation and value were then scaled by a sigmoidal function of the maximum overlap, such that positions with insignificant overlap appeared white.

\subsection{1D path integration}

The heteroassociative matrix $\mat{H}$ (\ref{eqn:H_matrix_1d}) is constructed using the derivative $\frac{\de \hv{x}}{\de p}$ and a bound dense bipolar vector $\hv{s}$. To avoid difficulties associated with differentiating the discontinuous position embedding function (\ref{eqn:x_embedding_func}), for the purpose of calculating gradients we replace it with a surrogate function with a finite gradient instead \cite{neftci_surrogate_2019}. Hence, $\hv{x}(p)$ is replaced with
\begin{equation}
x_i(p) = S\Big( \phi_i(p) - \frac{L-1}{2} \Big) \cdot S \Big( \phi_i(p) - \frac{L+1}{2} \Big)
\end{equation}
where $S(u) = 1/\big( 1 + \exp( -\beta u ) \big)$ is the logistic function with steepness parameter $\beta > 0$, serving as a smooth surrogate for the step function, and $\phi_i(p)$ is the modular phase given by
\begin{equation}
    \phi_i (p) = \mathrm{mod}_L \big( \omega_i p + \theta_i + \frac{L-1}{2} \big)
\end{equation}

We have effectively shifted the acceptance region to the middle of the interval $[0 ,L)$ to stop the gradient wrapping around the extremities. The steepness parameter $\beta$ we heuristically set to $\beta = 5$ in all experiments.

The weight matrix $\mat{W}$ is then given by the superposition $\mat{W} = \mat{A} + \mat{H}$. The vector $\hv{s}$ is generated to be block-constant, such that neurons in the same block have the same value, either $+1$ or $-1$ with equal probability. Hence, any expectation over the elements of $\hv{s}$ will be 0.
Due to the inclusion of this bound term, the expected contribution to the postsynaptic sum from the heteroassociative terms, in the absence of input, is also 0, since
\begin{equation}
\mat{H}\hv{x}(q) = c \int \frac{\de \tilde{\hv{x}}} { \de p} \underbrace{\big( \bar{\hv{x}}(p) \had \hv{s} \big) \cdot \hv{x}(q) \, }_{\expect{\ldots} \: = \:  0} \de p \: \approx \:  0
\end{equation}
thus the $\mat{H}$ terms do not contribute meaningfully to the dynamics.
When there is input to the network, which we model as a selective inhibition, the state update rule is
\begin{equation}
    \hv{z}_{t+1} = \mathrm{bWTA} \big[ \mat{W} (\hv{z}_t \land \hv{i}_t) \big]
\label{eqn:bwta_input}
\end{equation}
where $\land$ is an element-wise $\mathrm{AND}$ operation, implementing inhibition, and $\hv{i}_t \in \{ 0, 1\}^N$ is the input mask at time $t$ \cite{cotteret_distributed_2025}. When there is no input to the network, $\hv{i}_t$ is a vector of all ones, reducing (\ref{eqn:bwta_input}) to the input-free state update rule (\ref{eqn:bwta_noinput}).
If we apply an input to the network, which inhibit all neurons for which $s_i$ is negative, leaving only the components for which $s_i = 1$, then the postsynaptic sum becomes
\begin{equation}
\begin{split}
\mat{W}\big( & \hv{x}(q)  \land H(\hv{s}) \big) \\
& = \big( \mat{A} + \mat{H} \big) \big( \hv{x}(q)  \land H(\hv{s}) \big) \\
& = \int \Big( \bar{\hv{x}} \bar{\hv{x}}^\intercal + c \frac{\de \tilde{\hv{x}}}{ \de p} \underbrace{ \big( \bar{\hv{x}} \had \hv{s} \big)^\intercal \Big) \big(\hv{x}(q) \land H(\hv{s}) \big)}_{\expect{\ldots} \: \propto \: \bar{K}(p-q)} \de p \\
% & \approx \frac{N}{2L} \int \bar{\hv{x}}(p)\bar{K}(p-q)  + c \frac{\de \tilde{\hv{x}}}{ \de p}\bar{K}(p-q)  \de p \\
& \approx \frac{N}{2L} \int \Big( \bar{\hv{x}}(p)  + c \frac{\de \tilde{\hv{x}}}{ \de p} \Big) \bar{K}(p-q)  \de p \\
& \approx \frac{N}{2L}  \int \Big( \bar{\hv{x}}(p+c) + \mathcal{O}(c^2) \Big) \bar{K}(p-q)  \de p
\end{split}
\end{equation}
where we have applied the Heaviside step function $H(\cdot)$ to element-wise convert the bipolar vector $\hv{s}$ into a binary vector, which we use as input. This is because $H(\hv{s})$ is precisely the input which inhibits all neurons for which $s_i = -1$.
As a result, the active position state $q$ no longer projects maximally to $q$ as in the autoassociative case, but instead approximates the position state $q + c$, causing the network state to shift by $c$ on each time step.

\subsection{Sparse block vector binding}

The states to represent 2D positions are generated by independently generating two separate scalar position embeddings (\ref{eqn:x_embedding_func}), and then binding them together via a block-wise Local Circular Convolution (LCC) operation, denoted $*_\mathrm{LCC}$ \cite{frady_variable_2023}. Using the block and within-block indices $m$ and $n$ to describe the neuron index $i = n + Lm$, the LCC binding operation for $\hv{c} = \hv{a} *_{\mathrm{LCC}} \hv{b}$ is given as
\begin{equation}
    c_{n + Lm} = \sum_{k = 0}^{L-1} a_{\mathrm{mod}_L (n - k) + Lm} b_{k + Lm}
\label{eqn:lcc_definition}
\end{equation}
%
% To efficiently compute this however, we instead apply the convolution theorem within each block \cite{frady_variable_2023}.
To generate vectors representing higher-dimensional positions $\vec{g} \in \mathbb{R}^{D'}$, we similarly bind together $D'$ separate vector representations,
\begin{equation}
\hv{x} ( \vec{g}) = \hv{x}_1 (g_1 ) *_\mathrm{LCC}  \hv{x}_2 (g_2 ) \ldots *_\mathrm{LCC}  \hv{x}_{D^\prime} (g_{D^\prime} )
\label{eqn:lcc_bind_Ddim}
\end{equation}
where $\hv{x}_{1 \ldots D^\prime}$ are scalar position embedding functions (\ref{eqn:x_embedding_func}) with independently-sampled frequencies and offsets.
If additional restrictions are placed on the offsets $\theta_i$, then the vectors adhere to the relation $\hv{x}(p_1) *_\mathrm{LCC} \hv{x}(p_2) *_\mathrm{LCC}^{-1} \hv{x}(0)  \approx \hv{x}(p_1 + p_2) $, which allows arithmetic operations to be performed by direct manipulation of the distributed representations \cite{kymn_computing_2024, frady_computing_2022} (\ref{sec:SM_lcc}).

\subsection{2D plane attractor}
\label{sec:2d_plane_methods}

The autoassociative matrix is constructed as
\begin{equation}
\mat{A} = \iint_\text{plane} \bar{\hv{x}}(u, w) \bar{\hv{x}}(u, w)^\intercal  \de u \, \de w
\end{equation}
with the position representations generated according to
\begin{equation}
\bar{\hv{x}}(u, w) = \hv{x}_u(u) *_\mathrm{LCC} \hv{x}_w(w) - \frac{1}{L}
\end{equation}
where the two scalar position embedding functions $\hv{x}_{u/w}$ have independently-sampled embedding frequencies and offsets. The superimposed heteroassociative matrix, to embed a vector field $\vec{v}_\nu(\cdot) : \mathbb{R}^2 \rightarrow \mathbb{R}^2$, is given by
\begin{equation}
\mat{H}_\nu = \iint_\text{plane} \Big(\big( \vec{v}_\nu \cdot \nabla_{p} \big) \tilde{\hv{x}} \Big) (\bar{\hv{x}} \had \hv{s}_\nu)^\intercal \de u \, \de w
\end{equation}
where $\nabla_p = (\partial_u, \partial_w)^\intercal$ is a vector of partial derivative operators with respect to the position coordinates $\vec{p} = (u, w)^{\intercal}$, and $\hv{s}_\nu$ is a block-constant dense bipolar vector, analogous to the 1D integration case, but independently generated for each vector field $\nu$. 
The gradients $\nabla_{p} \,  \tilde{\hv{x}}$ are calculated by the method of surrogate gradients as earlier described, differentiating through the LCC binding operation, and are computed using JAX auto-differentiation functions \cite{bradbury_jax_2018}. The three imposed vector fields are shown in Fig. \ref{fig:integration_megafig}e, and the weights are given by the superposition
\begin{equation}
\mat{W} = \mat{A} + \sum_{\text{fields } \nu} \mat{H}_\nu   
\end{equation}
Identical to the 1D line attractor case, the individual vector fields $\vec{v}_\nu$ are triggered by inhibiting the neurons for which the corresponding $\hv{s}_\nu$ is negative.

\subsection{Curved manifold embeddings}
To embed curved manifolds with arbitrary topologies, we consider a smooth vector mapping $\vec{g}(\vec{p}) : \mathbb{R}^D \rightarrow \mathbb{R}^{D^\prime}$, with $D^\prime \geq D$ from our position coordinates $\vec{p}$ to a higher dimensional space, whose coordinates we embed into the attractor network with (\ref{eqn:lcc_bind_Ddim}).
The $\mat{A}$ and $\mat{H}$ matrices are constructed as previously described, integrating over the positions $\vec{p}$, except now we need to introduce additional state weighting terms to account for the non-uniformity of the $\vec{p} \rightarrow \vec{g}$ mapping.
The autoassociative matrix is now given by
\begin{equation}
\mat{A}_{\text{FO}} = \int \frac{1}{\sqrt{\det G }} \: \bar{\hv{x}}\big( \vec{g}(\vec{p}) \big) \bar{\hv{x}} \big( \vec{g}(\vec{p}) \big)^{\intercal} \de P
\end{equation}
where
\begin{equation}
\big[ G \big]_{ij} = \sum_{k=1}^{D^\prime} \frac{\partial g_k}{\partial p_i} \cdot \frac{\partial g_k}{\partial p_j} 
\end{equation}
This ensures that, to a first-order expansion of $\vec{g}$, the expected energies of all points far away from the edges of the manifold are equal.
% (\ref{sec:SM_stochproc}).
However, this does not account for expected energy differences due to manifold curvature, as those are stored in higher-order terms.
To account for these higher-order effects, we construct the weights in a two-step procedure.
First, we empirically measure the energy $\hat{E}$ for each state with respect to the first-order-corrected weight matrices, $\hat{E}(\vec{p}) = -\hv{x}^{ \intercal}\mat{A}_{\text{FO}} \hv{x}$, and then construct weight matrices inversely weighted by these energies. Thus the implemented matrix is given by
\begin{equation}
\mat{A} = \int \frac{1}{|\hat{E}| \sqrt{\det G }} \: \bar{\hv{x}} \bar{\hv{x}}^{ \intercal} \de P
\end{equation}

The superimposed heteroassociative matrices are similarly constructed with these weighting terms. 
\begin{equation}
\mat{H}_\nu = \int  \frac{1}{|\hat{E}| \sqrt{\det G }} \Big(\big( \mat{J}\vec{v}_\nu \cdot \nabla_{g} \big) \tilde{\hv{x}} \Big) (\bar{\hv{x}} \had \hv{s}_\nu)^\intercal \de P
\end{equation}
where $J_{ij} = \partial g_i / \partial p_j$ is a Jacobian, and the gradients of the position states $\nabla_{g} \: \tilde{\hv{x}}$ are with respect to $\vec{g}$, such that the vector fields $\vec{v}_\nu (\vec{p}) : \mathbb{R}^D \rightarrow \mathbb{R}^D$ are defined in terms of the original position coordinates $\vec{p}$.

\subsection{Comparison models}

Both comparison models from \textcite{zhang_representation_1996} and \textcite{kilpatrick_wandering_2013} have field equations that are continuous in space, but which we discretise into an equal number of positions $i = 1 \ldots N$ as neurons used in our network. Their dynamics are given by
\begin{equation}
    \tau \frac{\de  \hv{u}}{\de t} = -\hv{u} + \mat{W} f(\hv{u})
\end{equation}
where $\hv{u} \in \mathbb{R}^N$ is a vector of the the total synaptic currents to each neuron, $\tau > 0$ is a time constant, $\mat{W}$ are the recurrent weights, and $f(\cdot)$ is an element-wise current-to-firing-rate function. In the Zhang model, $f$ is a sigmoid function, and the weights are determined by calculating the convolution on $f(\hv{u})$ necessary to support stable bumps of predetermined shape \cite{zhang_representation_1996}. In the Kilpatrick \& Ermentrout model, $f(\hv{u}) = H(\hv{u} - \theta_\text{kp})$ is a Heaviside step function with threshold $\theta_\text{kp}$, and the weights are given by
\begin{equation}
    w_{ij} = \Big(1 + \sigma_\text{kp} \cos \Big( \frac{2 \pi}{N} n j \Big) \Big) \cos \Big(\frac{2 \pi}{N} (i - j) \Big)
\end{equation}
where $\sigma_\text{kp}$ and $n$ are the strength and frequency of the structured heterogeneity respectively \cite{kilpatrick_wandering_2013}. We used near-identical parameters to those given in each work (Table \ref{tab:SM_parameters}).
% , making only slight changes to ensure stability for our form of imposed nonidealities.
The estimated represented position $\hat{p}$ at any time was calculated by the firing-rate-weighted circular mean
\begin{equation}
\hat{p}(t) = \frac{1}{2 \pi} \mathrm{atan2}\Bigg[ \frac{\sum_i f(u_i) \cos\frac{2 \pi}{N} i }{\sum_i f(u_i)}, \frac{\sum_i \ldots \sin }{\sum_i f(u_i)} \Bigg] + \frac{1}{2}
\end{equation}
where we map the range $[0, 2\pi]$ to $[0,1]$ for comparison with our model. The continuous-time equations were simulated using Euler's method with a time step of $0.2 \tau$ seconds. Fixed synaptic nonidealities were added to $\mat{W}$ in both models as described for our model. Fast stochastic neural nonidealities were added in such a way to best allow comparison with our model. With probability $b$, each neuron's synaptic current $u_i$ was set to the highest or lowest current value across all neurons at that time. They were then clamped to these values for $2 \tau$ seconds, to allow the network to adjust to the errors. Afterwards, all the currents were allowed to freely evolve for $2 \tau$ seconds, again to allow the network to reach a new equilibrium, after which the cycle repeats. We used a slightly unrealistic time constant of $\tau = \qty{0.25}{\ms}$, such that \qty{1}{\ms} of noise in the comparison models was equal to 1 time step of noise in our model.

\section*{References}
\addcontentsline{toc}{section}{References}

\widowpenalty=10000
\clubpenalty=10000

\printbibliography[heading=none]

\end{refsection}

\begin{refsection}

% \onecolumn
% \newpage

\FloatBarrier

\section*{Acknowledgements}
\addcontentsline{toc}{section}{Acknowledgements}
We thank Mark Goldman for his help reviewing the quasi-continuous attractor literature, and Jake Pearson for creating the 3D rendered figures in Fig. \ref{fig:concept_fig}. This work has been supported by the NeuroPAC fellowship, an NSF AccelNet grant, and DFG projects NMVAC (432009531) and MemTDE (441959088).
The authors would like to acknowledge the financial support of the CogniGron research center and the Ubbo Emmius Funds (Univ. of Groningen).
We thank the Center for Information Technology of the University of Groningen for their support and for providing access to the Hábrók high performance computing cluster.

% \section*{Author contributions}
% \addcontentsline{toc}{section}{Author contributions}

% All authors contributed to writing the manuscript.

\section*{Competing interests}
The authors declare no competing interests.
 % \ref{sec:supp1}

\onecolumn

\newpage

\begin{center}
\textsc{Supplementary Material}
\end{center}

\renewcommand\thefigure{S\arabic{figure}}
\renewcommand\thetable{S\arabic{table}}
\renewcommand\thesection{S\arabic{section}}
\renewcommand{\theHfigure}{S\arabic{figure}}
\renewcommand\theHsection{S\arabic{section}}

\setcounter{figure}{0}
\setcounter{section}{0}
\setcounter{equation}{0}
\setcounter{page}{1}

\section{Kernel calculation}
\label{sec:SM_kernels}

The (normalised) kernel $K$ is given by probability that an active neuron in the representation of position $p$ is also active for position $p + \Delta p$,
\begin{equation}
    K(\Delta p ) = L \mathbb{E}_{\omega, \theta} [ x_i(p) x_i(p + \Delta p) ] = \mathbb{E}_{\omega, \theta} [x_i(p+\Delta p) | x_i(p) = 1]
\label{eqn:SM_kernel_def}
\end{equation}
The $L$ prefactor ensures that the inner product between a vector and itself is 1, i.e. $K(0) = 1$. We will henceforth drop the neuron index $i$, since the expected inner product is equal at every index.
Because the offsets $\theta \in [0,L)$ are uniformly distributed across the acceptance region of the neural activation function, and since we are conditioning on $x(p) = 1$, then (\ref{eqn:SM_kernel_def}) is equivalent to
\begin{equation}
\begin{split}
K(\Delta p) & = \mathbb{E}_{\omega} \Big[ \probP\big[ \mathrm{mod}_L (\omega \Delta p + t ) < 1 \big] \Big] \\
& = \mathbb{E}_\omega \Big[ \mathrm{ReLU}\big( 1 - \mathrm{mod}_L (\omega \Delta p) \big) + \mathrm{ReLU}\big(1 - \mathrm{mod}_L ( \omega \Delta p - L + 1)\big) \Big]
\end{split}
\end{equation}
where we introduce the uniform random variable $t \sim \mathcal{U}_{[0,1)}$ representing the distance to the edge of the acceptance region in either direction (positive or negative $\omega \Delta p$), and $\mathrm{ReLU}$ is a rectified-linear function.
We then want to write the above function as a Fourier series, periodic with period $L$. To make life easier, we will shift the domain of interest from $[0,L)$ to $[-L/2, L/2)]$. Then, the above expression is given by
\begin{equation}
\mathbb{E}_{\omega} \Big[ \mathrm{tri} \big( \mathrm{wrap}_{-\frac{L}{2}, \frac{L}{2}} (\omega \Delta p) \big) \Big]
\end{equation}
where $\mathrm{tri}$ is a triangle function and $\mathrm{wrap}_{-\frac{L}{2}, \frac{L}{2}} $ wraps its input to the desired range. The Fourier coefficients $C_n$ of the triangle function are easily calculated, giving
\begin{equation}
\begin{split}
K(\Delta p) & = \mathbb{E}_\omega \Big[ \sum_{n = -\infty}^{\infty} C_n e^{\frac{2 \pi i n }{L} \omega \Delta p} \Big] \\
& = \frac{1}{L} \sum_{n = -\infty}^{\infty}  \mathrm{sinc}^2 \Big( \frac{n}{L} \Big) \mathbb{E}_\omega \Big[ e^{\frac{2 \pi i n }{L} \omega \Delta p} \Big] \\
& = \frac{1}{L} \sum_{n = -\infty}^{\infty}  \mathrm{sinc}^2 \Big( \frac{n}{L} \Big)  \int_{-\infty}^\infty P(\omega) e^{\frac{2 \pi i n }{L} \omega \Delta p} \de \omega \\
& = \frac{1}{L} \sum_{n = -\infty}^{\infty}  \mathrm{sinc}^2 \Big( \frac{n}{L} \Big) \mathcal{F} \big[ P(\omega) \big]\Big(\frac{2 \pi  n}{L} \Delta p \Big) \\
& = \frac{1}{L} \sum_{n = -\infty}^{\infty}  \mathrm{sinc}^2 \Big( \frac{n}{L} \Big) K_{\text{RFF}} \Big(\frac{2 \pi n}{L} \Delta p \Big) \\
& = \frac{1}{L} + \frac{2}{L} \sum_{n = 1}^{\infty}  \mathrm{sinc}^2 \Big( \frac{n}{L} \Big) K_{\text{RFF}} \Big(\frac{2 \pi n}{L} \Delta p \Big)
\end{split}
\end{equation}
where $\mathrm{sinc}(u) = \sin (\pi u) /  \pi u$, and $K_\text{RFF} : \mathbb{R} \rightarrow [0,1]$ is the random Fourier feature kernel, corresponding to the phasor state embedding function $x_j(p) = \exp i( \omega_j p + \theta_j)$, with $\theta_j \sim \mathcal{U}_{[0, 2 \pi)}$ \cite{rahimi_random_2007,thomas_theoretical_2022, frady_computing_2022}. 
Note that in these equations we're treating $\omega$ as frequency, not an angular frequency.

As an example, if the frequencies are drawn from a rectangular distribution, whose Fourier transform is a $\mathrm{sinc}$ function, then
\begin{equation}
P(\omega) \propto \mathrm{rect}\Big(\frac{\omega}{4 \wma} \Big) \quad \longrightarrow \quad  K(\Delta p) = \frac{1}{L} + \frac{2}{L} \sum_{n=1}^{\infty} \mathrm{sinc}^2 \Big( \frac{n}{L} \Big) \mathrm{sinc} \Big( \frac{4n}{L} \wma \Delta p \Big)
\end{equation}
while for a Gaussian $P(\omega)$, whose Fourier transform is also Gaussian we have
\begin{equation}
P(\omega) \propto \exp \Big(-\omega^2 / (2 \sigma_\omega^2) \Big) \quad \longrightarrow \quad  K(\Delta p) = \frac{1}{L} + \frac{2}{L} \sum_{n=1}^{\infty} \mathrm{sinc}^2 \Big( \frac{n}{L} \Big) \exp \Big(-\frac{(2 \pi n \sigma_\omega)^2 \Delta p^2}{2 L^2} \Big)
\end{equation}
where $\sigma_\omega = \sqrt{\frac{\pi}{2}} \wma$ is the Gaussian's standard deviation.
The sampling distributions $P(\omega)$ and exact and measured kernels $K(\Delta p)$ are plotted in Fig. \ref{fig:integration_megafig}g,h with the first hundred $n$ terms calculated in each summation.

\section{Multidimensional kernels}
\label{sec:SM_kernel_multi}

\begin{figure}
\centering
\includegraphics[width=5in]{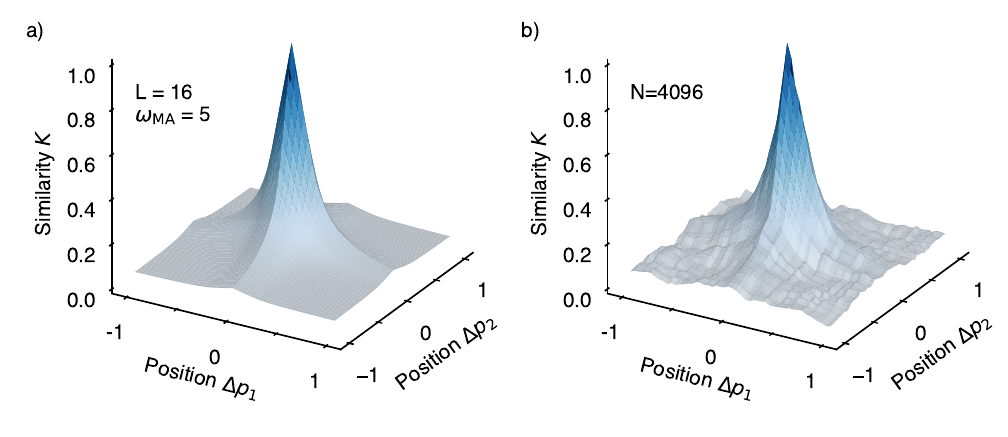}
\caption{Measured \textbf{(a)} and theoretical \textbf{(b)} 2D similarity kernels, with the embedding frequencies $\omega_i$ drawn from a rectangular distribution.}
\label{fig:SM_lcc_multidim}
\end{figure}

We wish to obtain a similar approximation for the similarity kernels of the multidimensional vector embeddings, in the 2D case defined by
\begin{equation}
\begin{split}
K(p_1 + \Delta p_1, p_2 + \Delta p_2) & = \frac{L}{N} \expect{\hv{x}(p_1, p_2) \cdot \hv{x}(p_1 + \Delta p_1, p_2 + \Delta p_2)} \\
& = L \mathbb{E} \big[  x_i (p_1, p_2) x_i (p_1 + \Delta p_1, p_2 + \Delta p_2) \big] \\
& = \mathbb{E} \big[ x_i (p_1 + \Delta p_1, p_2 + \Delta p_2)  |  x_i (p_1, p_2)  = 1 \big]    
\end{split}
\label{eqn:SM_multidim_kernel_def}
\end{equation}
for some index $i$. We consider the hypervector $\hv{c}(p_1, p_2)$ formed by the LCC binding operation of two hypervectors $\hv{a}(p_1)$ and $\hv{b}(p_2)$, i.e.
\begin{equation}
    \hv{c}(p_1, p_2) = \hv{a}(p_1) \LCC \hv{b}(p_2)
\label{eqn:SM_lcc_multidim_def}
\end{equation}
To calculate (\ref{eqn:SM_multidim_kernel_def}), we need to calculate the probability that an active index in $\hv{c}$ continues to be active when shifting in encoded position by $(\Delta p_1, \Delta p_2)$.
We use $A,B,C$ to denote the events that an active index in $\hv{a}, \hv{b}, \hv{c}$ remains active after this shift, respectively. Then, per the definition of LCC binding, we have
\begin{equation}
\begin{split}
\probP (C) & = \probP (C | A, B)\probP(A, B) + \probP(C | A, \text{ not } B) \probP (A, \text{ not } B) + \text{v.v.} + \probP (C | \text{ not } A, \text{ not } B) \\
& \approx 1 \cdot \probP(A) \probP(B) + 0 \cdot \big[ \probP (A) (1 - \probP(B)) + \text{v.v.} \big] + \frac{1}{L-1} \big( 1 - \probP(A) \big) \big( 1 - \probP(B) \big) \\
& = \frac{L}{L-1} \big[  \probP(A) \probP(B) - \frac{1}{L}\probP(A) - \frac{1}{L} \probP(B) + \frac{1}{L} \big]
\end{split}
\end{equation}
where on the second line we assume that, if the active indices in both $\hv{a}$ and $\hv{b}$ change, then there is a $1/(L-1)$ probability that the new indices combine to leave $\hv{c}$ unchanged. We then recognise that the probabilities $\probP(A)$ and $\probP(B)$ are precisely the 1D kernels we calculated earlier, giving us
\begin{equation}
K(\Delta p_1, \Delta p_2) = \frac{L}{L-1} \Big[ K_1(\Delta p_1) K_2(\Delta p_2) - \frac{1}{L} \big( K_1(\Delta p_1) + K_2(\Delta p_2)\big) + \frac{1}{L} \Big]
\label{eqn:SM_multidim_lcc_kernel_formula}
\end{equation}

This expression and an empirically-measured similarity kernel are shown in Fig. \ref{fig:SM_lcc_multidim}. To obtain an expression for higher-dimensional kernels, we can iteratively apply (\ref{eqn:SM_multidim_lcc_kernel_formula}), yielding in the 3D case for example
\begin{equation}
K(\Delta p_1,  \Delta p_2, \Delta p_3) = \frac{1}{(L-1)^2} \Big[ L^2 K_1 K_2 K_3 - L \big( K_1 K_2 + K_2 K_3 + K_3 K_1 \big) + (K_1 + K_2 + K_3) + (L-2) \Big]
\end{equation}
where $K_s = K_s( \Delta p_s), \: s = \{1,2,3\}$ are the 1D kernels in each dimension.

\newpage

\section{Noisy inner-products as a stochastic process}
\label{sec:SM_stochproc}

Since we know the expected inner product between position vectors, we can rewrite inner products as
\begin{equation}
     \hv{x}(p) \cdot \hv{x}(p + \Delta p) / M = K(\Delta p) + \epsilon(p, \Delta p)
\end{equation}
where $\epsilon$ is the deviation of the inner products from their expected values. Although we can apply CLT-based techniques to estimate the magnitude of $\epsilon$ (or apply more rigorous methods \cite{thomas_theoretical_2022}), we don't have a way of estimating how $\epsilon$ varies as the positions are varied. If $p$ is changed by a very small amount, we would expect $\epsilon$ to be approximately constant, while for a large change in $p$, we expect that $\epsilon$ would be resampled from some underlying stationary distribution.
Understanding the dynamics of $\epsilon$ is necessary to derive properties of the energy landscape along the line attractor, principally the length scale of its heterogeneity.

We will calculate the overlap between a binary sparse (non-block) hypervector $\hv{x}(p)$ of dimension $N$, with average number of nonzero components $M = \frac{N}{L}$, and a vector of all ones $\hv{1}$. Applying this calculation to two sparse block vectors will require only that some of the quantities be relabelled at the end.
We split the domain of possible $\omega$ values into equally-sized bins of width $\eta$, indexed by $k$. Assuming that $\eta$ is small enough, we can approximate neurons in the same bin $k$ as having the same frequency $\omega_k$. We denote the number of neurons in bin $k$ by $n_k$, with expected occupancy $\langle n_k \rangle = N P(\omega_k) \eta $. We denote the contribution to the inner product $\hv{x}(p)\cdot \hv{1}$ from bin $k$ as $X_k \in \{0, \ldots, n_k \}$, and consider how $X_k$ varies as $p$ is changed by a small amount $p \rightarrow p + \delta p$.
Since $\delta p$ is small, the only possibilities for change are:
\begin{enumerate}
    \item a component for which $x_i(p) = 1$ turns off, $x_i(p + \delta p) = 0$.
    \item a component for which $x_i(p) = 0$ turns on, $x_i(p + \delta p) = 1$.
\end{enumerate}
The change in overlap $\delta X_k$ is thus given by
\begin{equation}
\begin{split}
    \delta X_k   & =  \delta X_k^+ - \delta X_k^- \\
    \delta X_k^+ & \sim B \big( n_k - X_k, \frac{\omega_k \, \delta p}{L-1} \big) \\
    \delta X_k^- & \sim  B \big( X_k, \frac{\omega_k \, \delta p}{1} \big )
\end{split}
\end{equation}
where $B(n, p)$ is the binomial distribution, and $\delta X_k^+$ and $\delta X_k^-$ are the number of neurons that switch on and off respectively.

\subsection{Ornstein-Uhlenbeck approximation}

\begin{figure}
    \centering
    \includegraphics[width=2.5in]{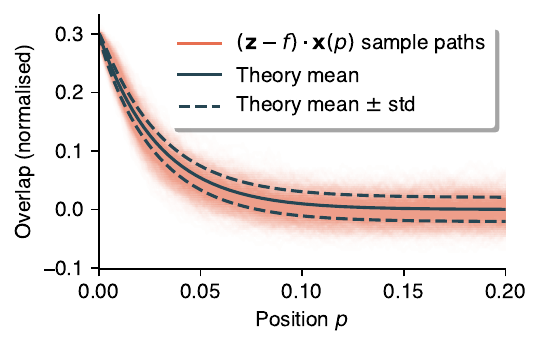}
    \caption{10,000 sample paths of the mean-adjusted overlap error between two sparse block code hypervectors, as the represented position $p$ for one of them is varied. Each path represents a resampling of the offsets and frequencies. The static hypervector $\hv{z}$ was constructed to have an initial overlap error of $0.3$ for each path. Overlain is the theoretical distribution, derived from the OU approximation.}
    \label{fig:SM_stochproc_overlaps}
\end{figure}

In order to make the dynamics of $X_k$ amenable to mathematical manipulation, we now try to approximate its dynamics with a Gaussian noise process $\de W$, which has variance $\expect{\de W^2} = \de p$ and mean $\expect{\de W} = 0$. We can thus construct a stochastic differential equation (SDE) which has the same mean and variance for a small step in $p$, given by
%
% We can now try to approximate this using a Gaussian noise process $\de W$, which has variance $\langle \de W^2 \rangle = \de p$, by constructing a stochastic differential equation that has the same mean and variance for a small step in $p$. It is given by
%
\begin{equation}
\begin{split}
    \de X_k & = \Big[(n_k-X_k) \frac{\omega_k}{L-1} - X_k \omega_k \Big] \de p + \sqrt{X_k \omega_k + (n_k-X_k) \frac{\omega_k}{L-1}} \de W \\
    & = \frac{\omega_k}{L-1} \big[ n_k - X_k L \big] \de p + \sqrt{\frac{\omega_k}{L-1} \big[ n_k + X_k (L-2) \big] } \de W
\end{split}
\end{equation}
where we have used that the variance of a Binomial distribution is approximately equal to the mean for very small probabilities. This expression gives us the dynamics of the contribution to the overlap from bin $k$. To obtain an expression for the total overlap, we thus sum the contributions from all bins $k$ and take the limit $\eta \rightarrow 0$. We will make the assumption that the fraction of 1-valued neurons in each bin is equal to the total fraction of 1-valued neurons across all bins, i.e.
\begin{equation}
    \frac{X_k}{n_k} = \frac{X}{N} \quad \rightarrow \quad X_k = \frac{n_k}{N} X = P(\omega_k) \eta X
\end{equation}
where $X = \sum_k X_k$, $X \in \{0, \ldots, N\}$ is the total inner product.
Then, summing contributions from each bin $k$ and taking limits, we get 
\begin{equation}
\begin{split}
\de X & = \lim_{\eta \rightarrow 0} \sum_k \de X_k \\
& = \frac{1}{L-1}\lim_{\eta \rightarrow 0} \Big[ \sum_k \omega_k n_k  - L \sum_k \omega_k X_k \Big] \de p + \sqrt{ \frac{1}{L-1} \lim_{\eta \rightarrow 0}  \Big[\sum_k \omega_k n_k  + (L-2) \sum_k \omega_k X_k \Big] } \de W \\
& = \frac{1}{L-1} \Big[ N \wma - LX \wma \Big] \de p + \sqrt{\frac{1}{L-1} \Big[ N \wma + (L-2) X \wma \Big] } \de W \\
& = \frac{\wma L}{L-1} \Big[ \frac{N}{L} - X \Big] \de p + \sqrt{\frac{\wma }{L-1} \Big[ N + (L-2) X \Big] } \de W
\end{split}
\end{equation}
where $\wma = \int | \omega | P(\omega) \de \omega$ is the mean absolute embedding frequency. For a Gaussian $P(\omega)$, this is $\sqrt{\frac{2}{\pi}} \approx 0.8$ times the standard deviation.
This expression can be simplified further. Since we know that independent SBC hypervectors will have an inner product tightly-concentrated around $N/L$, with standard deviation $\sqrt{N/L}$, we can replace $X$ in the noise term with its mean value. This is of course valid only for large $N$ (as is the bedrock of this derivation), and we are presuming that slight fluctuations in the magnitude of the noise term can be neglected. This gives
\begin{equation}
\begin{split}
\de X & \approx \frac{\wma L}{L-1} \Big[ \frac{N}{L} - X \Big] \de p + \sqrt{\frac{\wma }{L-1} \Big[ 2N -2\frac{N}{L} \Big] } \de W \\
& =  \frac{\wma L}{L-1} \Big[ \frac{N}{L} - X \Big] \de p + \sqrt{\frac{2 N}{L} \wma } \de W \\
% & \approx  \wma \Big[ M - X \Big] \de p + \sqrt{2 M \wma } \de W \\
\end{split}
\end{equation}
This SDE describes the inner product between a vector of all ones, and a sparse (not necessarily block) hypervector. To obtain the SDE for the inner product between two sparse block vectors, we need only to replace $N$ and $M$ with $N/L$ and $M/L$ respectively, to reflect that the SBC hypervector is effectively just a vector of all ones in $1/L$ of its sites. This gives
\begin{equation}
\de X = \wma \frac{L}{L-1} \Bigg[ \frac{N}{L^2} - X \Bigg] \de p + \sqrt{\frac{2N}{L^2} \wma } \de W
\label{eqn:SM_dX_with_offset}
\end{equation}
Hence, the inner-product between two sparse block hypervectors, as the position of one of them is varied, follows a mean-reverting noisy process. This has a steady-state solution with mean $N/L^2$ and variance $\frac{N}{L^2} \frac{L-1}{L}$, as is expected of the inner product between two independent SBC hypervectors.
What will be more useful for subsequent calculations however, is the inner product between two SBC hypervectors with the mean subtracted. For this, we redefine $X(p)$ as
\begin{equation}
X(p) := (\hv{z} - f)\cdot \hv{x}(p)
\end{equation}
This of course amounts only to subtracting the offset from (\ref{eqn:SM_dX_with_offset}), giving
\begin{equation}
\begin{split}
\de X & = -\wma \frac{L}{L-1}X \de p + \sqrt{\frac{2N}{L^2} \wma } \de W \\
& = \vphantom{\frac{L}{L}}-\gamma X \de p + g \de W
\end{split}
\label{eqn:SM_OU}
\end{equation}
where we define $\gamma = \wma \frac{L}{L-1}$ and $g = \sqrt{2N\wma / L^2}$. This is the famed Ornstein-Uhlenbeck (OU) equation, originally describing the velocity of a pollen grain under the bombardment of air particles \cite{uhlenbeck_theory_1930, jacobs_stochastic_2010}, but here modelling the residual of the inner product between SBC vectors. This has solution
\begin{equation}
X(p) = X_0 e^{-\gamma p} + g \int_0^p e^{\gamma (p^\prime - p)} \de W^\prime
\label{eqn:SM_OU_soln}
\end{equation}
where $X_0$ is the value of $X$ at $p = 0$. In Fig. \ref{fig:SM_stochproc_overlaps} we compare empirically-measured overlaps and the distribution predicted by (\ref{eqn:SM_OU_soln}).

\subsection{Residual autocovariance}

\begin{figure}
\centering
\includegraphics[width=2in]{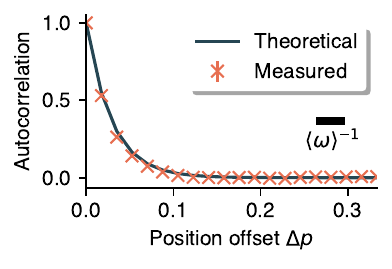}
\caption{Autocorrelation for the cross-talk noise (deviation from the expected kernel $K$) for different offsets between position states. Specifically, this is looking at $\mathrm{Corr}\big[\bar{\hv{z}} \cdot \hv{x}(p), \bar{\hv{z}} \cdot \hv{x}(p+\Delta p) \big](\Delta p)$. Overlain is the theoretical autocorrelation from the OU approximation, which is a decaying exponential with length scale $\wma^{-1}$ (shown by scale).}
\label{fig:autocorr_overlaps}
\end{figure}

The autocovariance of the residual, $\expect{X(p) X(p+\Delta p)}$ will be needed to calculate the autocorrelation of the energies. Although this calculation can be lifted from other sources \cite{jacobs_stochastic_2010, uhlenbeck_theory_1930}, for completeness it is given here by
\begin{equation}
\begin{split}
\expect{X(p)X(p+\Delta p)} & = \iint X_1 X_0 \mathbb{P}[X_1(p + \Delta p)| X_0(p)] \mathbb{P}[X_0(p)] \de X_0 \de X_1 \\
& = \iint \frac{1}{2 \pi \sigma(X_0) \sigma_0}X_1 X_0 e^{-(X_1-\mu(X_0))^2/ (2 \sigma(X_0)^2)} e^{-X_0^2 / (2 \sigma_0^2)} \de X_0 \de X_1 \\
& = \int \frac{1}{\sqrt{2 \pi} \sigma_0} X_0 \mu(X_0) e^{X_0^2 /2 \sigma_0^2} \de X_0 \\
& = \int \frac{1}{\sqrt{2 \pi} \sigma_0 } X_0 \big( X_0e^{-\gamma \Delta p} \big) e^{X_0^2 /2 \sigma_0^2} \de X_0 \\
& = \sigma_0^2 e^{-\gamma \Delta p}
\label{eqn:XX_overlaps}
\end{split}
\end{equation}
where $\sigma_0^2 = g^2/(2\gamma)$ is the steady-state (unconditioned) variance of $X$. This is plotted in Fig. \ref{fig:autocorr_overlaps}. From this we can conclude that the length scale over which $X$ ``forgets'' its value is approximately $1/ \gamma \approx 1/\wma$.

\subsection{Energy statistics}

The energy of a state $\hv{x}(p)$, if the weight matrix consists only of the symmetric matrix $\mat{A}$, is given by
\begin{equation}
\begin{split}
E(\hv{x}(p)) & = - \hv{x}(p)\T \mat{A} \hv{x}(p) = - \hv{x}(p)\T \int_0^P \bar{\hv{x}}(p^\prime) \bar{\hv{x}}(p^\prime)\T \de p^\prime \hv{x}(p) = \int_0^P \big( \bar{\hv{x}}(p^\prime) \cdot \hv{x}(p) \big)^2 \de p^\prime
\end{split}
\label{eqn:SM_E_int_def}
\end{equation}
where w.l.o.g. we set $P = \qty{1}{\metre}$, such that the line attractor represents positions between 0 and 1 metres, since increasing $P$ is equivalent to increasing the frequencies $\omega$. Because the energy consists of inner products between position states, $E(p)$ will also follow a stochastic process. To calculate the statistics of $E$, we will make the quite heavy-handed assumption that the residual error between two position states is well-modelled by (\ref{eqn:SM_OU}), and critically is unaffected by the value of $K$.
If one considers the residual from two very-distant states, this is a reasonable assumption, since then the states are almost completely independent, and $X$ may be sampled from its steady-state distribution. But, for two proximal states, this is obviously not true, since the error $\epsilon = \hv{x}(p) \cdot \hv{x}(p) - \frac{N}{L} K(p - p) = \frac{N}{L} - \frac{N}{L} = 0$ will be exactly 0 for identical states. We will thus be overestimating the error between vectors representing nearby positions.
Expanding (\ref{eqn:SM_E_int_def}) we get
\begin{equation}
\begin{split}
E(\hv{x}(p)) & \approx \int_0^P \big(  M  \bar{K}(p^\prime - p) +  \bar{\hv{z}}(p^\prime) \cdot \hv{x}(p) \big)^2 \de p^\prime \\
& = \int_0^P  M^2 \bar{K}^2(p^\prime - p) +2 M  \bar{K}(p^\prime - p) \big( \bar{\hv{z}}(p^\prime) \cdot \hv{x}(p) \big)  + \big( \bar{\hv{z}}(p^\prime) \cdot \hv{x}(p) \big)^2  \de p^\prime
\end{split}
\end{equation}
where $\overline{\hv{z}}(p')$ is a fictitious distant position representation vector, which enforces our above mentioned assumption by giving rise to errors with dynamics given by the OU approximation (\ref{eqn:SM_OU}).

To calculate the energy variance, we will need the autocorrelation function of $X^2(p) = \big( \bar{\hv{z}} \cdot \hv{x}(p)\big)^2$. We thus need to calculate
\begin{equation}
\begin{split}
\expect{(X_0^2 (p) - \sigma_0^2)( X^2(p+\Delta p) - \sigma_0^2)} & = \iint \big( X_0(p)^2 - \sigma_0^2 \big) \big( X(p+\Delta p) - \sigma_0^2 \big) \mathbb{P}[X_0(p), X(p+ \Delta p)] \de X_0 \de X \\
% & = \iint \big( X_0(p)^2 - \sigma_0^2 \big) \big( X(p+\Delta p) - \sigma_0^2 \big) \frac{1}{\sigma \sqrt{2 \pi}} e^{-(X-\mu)^2 / (2 \sigma^2)} \frac{1}{\sigma_0 \sqrt{2 \pi}} e^{-X_0^2/(2 \sigma_0^2)} \de X_0 \de X \\
& = \iint \big( X_0(p)^2 - \sigma_0^2 \big) \big( X(p+\Delta p) - \sigma_0^2 \big) \frac{1}{\sigma \sigma_0 2 \pi} e^{-(X-\mu)^2 / (2 \sigma^2)} e^{-X_0^2/(2 \sigma_0^2)} \de X_0 \de X \\
& = \frac{1}{2 \pi \sigma \sigma_0} \int \big( X_0^2 - \sigma_0^2 \big) e^{-X_0^2 / (2 \sigma_0^2)} \Bigg( \int_{-\infty}^\infty  (X^2 - \sigma_0^2) e^{-(X-\mu)^2 / (2 \sigma^2)} \de X \Bigg) \de X_0 \\
& = \frac{1}{\sigma_0 \sqrt{2 \pi}} \int \big( X_0^2 - \sigma_0^2 \big) e^{-X_0^2 / (2 \sigma_0^2)} (\sigma^2 + \mu^2 - \sigma_0^2) \de X_0 \\
\intertext{then subbing in the expressions for the mean $\mu(X_0)$ and standard deviation $\sigma(X_0)$ from (\ref{eqn:SM_OU_soln}),}
& = \frac{1}{\sigma_0 \sqrt{2 \pi}} \int \big( X_0^2 - \sigma_0^2 \big) \big( \sigma_0^2 \big[ 1 - e^{-2\gamma \Delta p} \big] + X_0^2 e^{-2\gamma \Delta p} - \sigma_0^2 \big) e^{-X_0^2 / (2 \sigma_0^2)} \de X_0 \\
& = \frac{1}{\sigma_0 \sqrt{2 \pi}} e^{-2 \gamma \Delta p} \int \big( X_0^4 + - 2\sigma_0^2 X_0^2 + \sigma_0^4 \big)  e^{-X_0^2 / (2 \sigma_0^2)}  \de X_0 \\
& = 2 \sigma_0^4 e^{-2 \gamma \Delta p} \\
& = 2 \sigma_0^4 e^{-2 \wma \frac{L}{L-1} \Delta p}
\label{eqn:autocorr_X2}
\end{split}
\end{equation}
which will later be useful. This says that the noisy contribution to the energy of unrelated (far separated in $p$) states has a forgetting length-scale of approximately $(2 \wma)^{-1}$.

The mean energy is similarly calculated as
\begin{equation}
\begin{split}
\expect{E(p)} & = \int_0^P  \expect{ M^2 \bar{K}^2(p^\prime - p)} -2 M  \expect{\bar{K}(p^\prime - p) \big( \bar{\hv{z}}(p^\prime) \cdot \hv{x}(p) \big)} + \expect{\big( \bar{\hv{z}}(p^\prime) \cdot \hv{x}(p) \big)^2}  \de p^\prime \\
& \approx M^2 \int_{-\infty}^{\infty} \bar{K}^2(p^\prime) \de p^\prime - 2M \int_{-\infty}^\infty \expect{K(p^\prime) X(p^\prime)} \de p^\prime + \int_0^P \expect{X^2 (p^\prime)} \de p^\prime \\
& \approx M^2 \frac{1}{\wma}\int_{-\infty}^{\infty} \bar{K}^2(x / \wma) \de x - 2M \int_{-\infty}^\infty \expect{K(p^\prime)} \underbrace{\expect{X(p^\prime)}}_{0} \de p^\prime + \int_0^P \sigma_0^2 \de p^\prime \\
& = M^2 \frac{\kappa_2}{\wma} + \sigma_0^2P \\
& = \frac{N^2}{L^2} \frac{\kappa_2}{\wma}  + \frac{N}{L^2} \frac{L}{L-1} P
\end{split}
\label{eqn:SM_mean_E}
\end{equation}
where we introduce the integral $\kappa_2 = \int_{-\infty}^\infty \bar{K}^2(x / \wma) \de x \approx 1$, which is independent of $\wma$ due to the the kernel's width scaling with $\wma$ too (\ref{sec:SM_kernels}). We neglect edge effects by assuming that $p$ is far from the integration bounds. Since $\bar{K} \approx 0$ for distances greater than $1/\wma$, on the second line we extended the integration bounds to infinity.
Assuming $L \gg 1$ (states are sparse), we can write (\ref{eqn:SM_mean_E}) as
\begin{equation}
\expect{E(p)} \approx \frac{N^2}{L^2 \wma} \Big(\kappa_2 + \frac{P \wma}{N}  \Big)
\end{equation}
which clearly splits the energy into a contribution from the kernel overlap (independent of the length of the attractor), and a contribution from the nonzero cross-talk overlaps (proportional to the length of the stored attractor $P \wma$). Note that off-manifold states will thus have a mean energy given by the rightmost term.

We are now in a position to calculate the autocorrelation length scale and variance of the energy. We first calculate $\big \langle E(\hv{x}(p)) E( \hv{x}(p + \Delta p)) \big \rangle$, which, keeping only terms with even powers of $X$, is given by
\begin{equation}
\begin{split}
\centering
\Bigg \langle \int_0^P \int_0^P \mathcolorbox{yellow}{M^4 \bar{K}^2 (p' - p ) \bar{K}^2 (p'' - p - \Delta p)} &  + \mathcolorbox{cyan}{4 M^2 \bar{K}(p' - p) \bar{K}(p'' -p - \Delta p) X(p',p) X(p'', p+\Delta p)} \\
& \hspace{-4em} + \mathcolorbox{pink}{2M^2 \bar{K}^2(p' -p) X^2(p'', p + \Delta p)} + \mathcolorbox{lime}{X^2(p', p) X^2(p'', p + \Delta p)} \de p' \de p'' \Bigg \rangle
\end{split}
\end{equation}
\textit{Divide et impera:}

\begin{figure}
\centering
\includegraphics[width=0.6\linewidth]{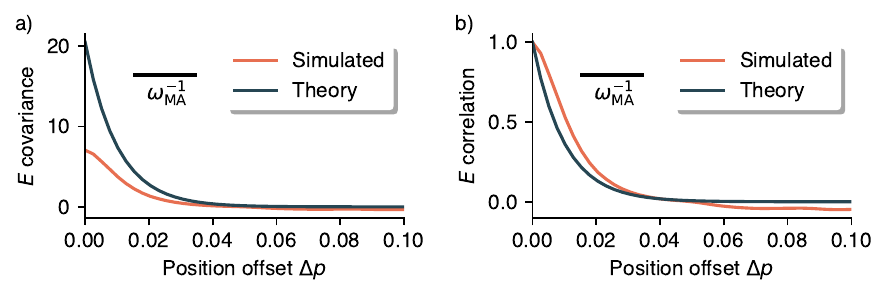}
\caption{The \textbf{a)} autocovariance and \textbf{b)} autocorrelation of the energy on the line attractor, for different offsets $\Delta p$. The theoretical estimate is given by (\ref{eqn:SM_covariance}). The discrepancy between the magnitude of the theoretical and simulated covariances is most likely due to the quite heavy-handed assumption, that the inner product error is independent of $\Delta p$. When the variance is divided out, the simulated correlations show good agreement with the theory, in particular the length scale over which the correlation decays.}
\label{fig:correlation_plot}
\end{figure}

The yellow terms are easily calculated. Assuming that $p$ is ``far'' from the edges of the line attractor, meaning many multiples of $\wma^{-1}$, we can safely extend the integration limits to $\pm \infty$. Then, it is given as
\begin{equation}
\begin{split}
    % \mathcolorbox{yellow}{\int_0^P \int_0^P M^4 \bar{K}^2 (p' - p ) \bar{K}^2 (p'' - p - \Delta p) \de p' \de p''}
     \colorbox{yellow}{$\ldots \vphantom{\int}$} & \approx \int_{-\infty}^\infty \int_{-\infty}^\infty M^4 \bar{K}^2 (p' - p ) \bar{K}^2 (p'' - p - \Delta p) \de p' \de p'' \\
    & = M^4 \int_{-\infty}^\infty \int_{-\infty}^\infty \bar{K}^2 (p' ) \bar{K}^2 (p'') \de p' \de p'' \\
    & = M^4 \Bigg[ \int_{-\infty}^\infty\bar{K}^2 (p' ) \de p' \Bigg]^2 \\
    & = M^4 \frac{1}{\wma^2 }\Bigg[ \int_{-\infty}^\infty \bar{K}^2 \Big( \frac{u}{\wma} \Big) \de u \Bigg]^2 \\
    & = \mathcolorbox{yellow}{\frac{M^4}{\wma^2} \kappa_2^2}
\end{split}
\end{equation}
where $\kappa_2 \approx 1$ is as previously defined. The blue terms are simplified by applying our assumption that the errors are independent of the kernel value. Note that this means we will be overestimating the variance contributed to this term, since we know that the error is reduced around $K(p\approx 0)$. Regardless,
\begin{equation}
\begin{split}
\qquad \qquad \colorbox{cyan}{$\ldots \vphantom{\int K^2}$} & = 4M^2 \int_0^P \int_0^P \bar{K}(p' -p) \bar{K}(p'' -p - \Delta p) \expect{X(p',p)X(p'', p + \Delta p)}  \de p' \de p'' \\
& \approx 4M^2 \int_{-\infty}^\infty \int_{-\infty}^\infty \bar{K}(p') \bar{K}(p'' - \Delta p) \expect{X(p',0)X(p'', \Delta p)}  \de p' \de p'' \\
\intertext{where we have assumed again that $p$ is far from the edges. We then assume that our expression for the autocorrelation (\ref{eqn:XX_overlaps}) can be applied for either of the states varying independently}
& \approx 4M^2 \sigma_0^2 \int_{-\infty}^\infty \int_{-\infty}^\infty\bar{K}(p')\bar{K}(p''- \Delta p) e^{- \gamma \Delta p} e^{- \gamma | p' - p''|} \de p' \de p'' \\
& = 4M^2 \sigma_0^2  e^{- \gamma \Delta p} \int_{-\infty}^\infty \int_{-\infty}^\infty \bar{K}(p') \bar{K}(p'') e^{- \gamma| p' - p'' - \Delta p|} \de p ' \de p'' \\
& =  \frac{4M^2 \sigma_0^2}{\wma^2}  e^{- \gamma \Delta p} \int_{-\infty}^\infty \int_{-\infty}^\infty \bar{K}\big(\frac{u}{\wma} \big) \bar{K} \big( \frac{v}{\wma} \big) \exp \Big[- \frac{\gamma}{\wma} | u - v - \wma \Delta p | \Big] \de u \de v \\
\intertext{where we have changed variables such that the kernels $\bar{K}$ are independent of $\wma$, and are nonzero only within $\approx 1$ distance of 0. Evaluating this integral is made difficult by the exponential absolute expression. But, given that the the kernels will be nonzero only for small $u$ and $v$, we can reasonably approximate the exponential to the value at $u = v = 0$,}
& \approx \frac{4M^2 \sigma_0^2}{\wma^2}  e^{- \gamma \Delta p} \int_{-\infty}^\infty \int_{-\infty}^\infty \bar{K}\big(\frac{u}{\wma} \big) \bar{K} \big( \frac{v}{\wma} \big) e^{-\gamma \Delta p} \de u \de v \\
& =  \mathcolorbox{cyan}{\frac{4M^2 \sigma_0^2}{\wma^2}  e^{- 2 \gamma \Delta p} \kappa_1^2}
\end{split}
\end{equation}
where we define $\kappa_1 = \int_{-\infty}^{\infty} \bar{K}(u / \wma) \de u \approx 1$. The pink terms are:
\begin{equation}
\begin{split}
\colorbox{pink}{$\ldots \vphantom{\int}$} & = \int_0^P \int_0^P 2 M^2 \bar{K}^2(p' - p) \expect{X^2(p'', p + \Delta p)} \de p' \de p'' \\
& \approx  2M^2  \int_{-\infty}^{\infty}  \int_{-\infty}^{\infty} K^2(p') \sigma_0^2 \de p' \de p'' \\
& = \mathcolorbox{pink}{\frac{2M^2 \sigma_0^2 P}{\wma} \kappa_2}
\end{split}
\end{equation}
%
% \newpage
And lastly we have the green terms, where again we will assume that our expression for the autocorrelation (\ref{eqn:autocorr_X2}) can be applied for both arguments independently,
\begin{equation}
\begin{split}
\hspace{8em} \colorbox{lime}{$\ldots \vphantom{\int}$} & = \int_0^P \int_0^P \big \langle X^2(p', p) X^2(p'', p + \Delta p) \big \rangle \de p' \de p'' \\
& \approx \int_0^P \int_0^P 2 \sigma_0^4 e^{-2\gamma \Delta p} e^{-2\gamma |p' - p''|} + \sigma_0^4 \:  \de p' \de p'' \\
& = P^2 \sigma_0^4 + 2 \sigma_0^4 e^{- 2 \gamma \Delta p} \int_0^P \int_0^P e^{-2\gamma |p' - p''|}  \de p' \de p'' \\
\intertext{change variables to $u = \frac{1}{2}(p' + p'')$ and $v = \frac{1}{2}(p' - p'')$}
& =  P^2 \sigma_0^4 + 2 \sigma_0^4 e^{- 2 \gamma \Delta p} \int_{u = 0}^{u = P} \int_{v = -|u - \frac{P}{2}| - \frac{P}{2}}^{v = |u - \frac{P}{2}| + \frac{P}{2} } e^{-4\gamma |v|}  \de v \de u \\
\intertext{which due to the integrand being $\approx 0$ for large $v$, and ignoring the behaviour at the extremities, can be simplified to }
& \approx  P^2 \sigma_0^4 + 2 \sigma_0^4 e^{- 2 \gamma \Delta p} \int_{u = 0}^{u = P} \int_{-\infty}^\infty  e^{-4\gamma |v|} \de v \de u \\
% & =  P^2 \sigma_0^4 + \sigma_0^4 e^{- 2 \gamma \Delta p} P \frac{1}{ \gamma} \\
& = \mathcolorbox{lime}{P^2 \sigma_0^4 + \frac{P \sigma_0^4}{\gamma} e^{- 2 \gamma \Delta p}}
\end{split}
\end{equation}
We can now combine these technicolor results to obtain the energy covariance, given by
\begin{equation}
\begin{split}
\hspace{6em} \mathrm{Cov}\big[ E(\hv{x}(p)), E(\hv{x}(p + \Delta p)) \big] & = \big \langle E(\hv{x}(p)) E( \hv{x}(p + \Delta p)) \big \rangle - \expect{E}^2 \\
& = \big \langle E(\hv{x}(p)) E( \hv{x}(p + \Delta p)) \big \rangle - \Big[ M^2 \frac{\kappa_2}{\wma} + \sigma_0^2 P \Big]^2 \\
& = \Big[ \frac{P \sigma_0^4}{\gamma} + \frac{4M^2 \sigma_0^2}{\wma^2}  \kappa_1^2 \Big] e^{- 2 \gamma \Delta p} \\
& = \frac{N^2}{L^4} \Big( \frac{L-1}{L} \Big) \frac{P}{\wma} \Bigg[ \Big( \frac{L-1}{L} \Big)^2 + \frac{4 N \kappa_1^2}{P \wma} \Bigg]e^{-2 \wma \frac{L}{L-1} \Delta p} \\
\intertext{which for very sparse states $L \gg 1$ simplifies to} 
& \approx \frac{N^2}{L^4} \frac{P}{\wma} \Bigg[1 + \frac{4 N \kappa_1^2}{P \wma} \Bigg]e^{- 2 \wma \Delta p}
\label{eqn:SM_covariance}
\end{split}
\end{equation}
where, the most important thing for us is that the heterogeneity of the energy surface has an autocorrelation length scale of approximately $(2\wma)^{-1}$. For a Gaussian $P(\omega)$, we can numerically integrate our kernel approximation to obtain the estimate $\kappa_1^2 \approx 1.5$. The theoretical estimate for the correlation and the simulated values are shown in Fig. \ref{fig:correlation_plot}. Although the estimate is off by a factor of 2 or so, there is good agreement with the length scale of the dependence upon $\Delta p$. 

\newpage

\FloatBarrier

\newpage

\section{LCC binding relation}
\label{sec:SM_lcc}

\begin{figure}
\centering
\includegraphics[width=5in]{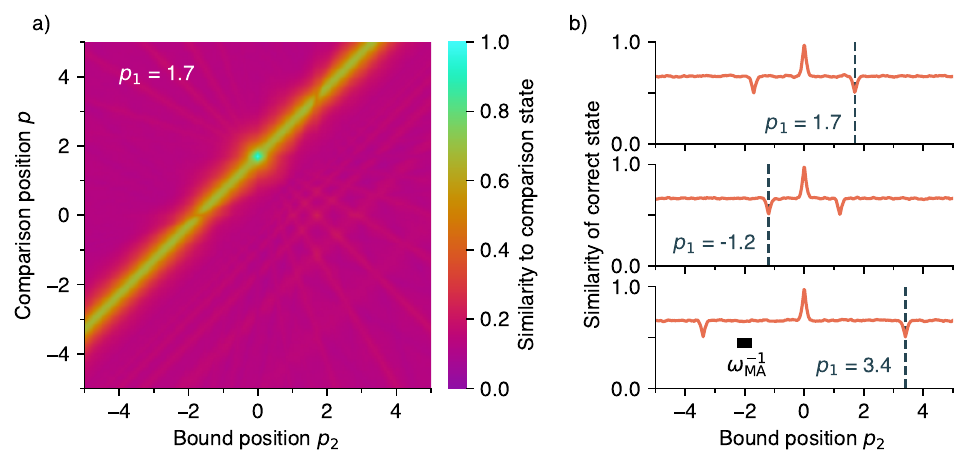}
\caption{LCC binding with sparse block code hypervectors implements addition in the encoded variables. \textbf{a)} The normalised similarity between the bound hypervector $\hv{x}(p_1) \LCC \hv{x}(p_2) \LCC^{-1} \hv{x}(0)$ for many values of $p_2$, and a comparison hypervector $\hv{x}(p)$. Here $p_1 = 1.3$, and the line of maximum overlap follows $p = 1.3 + p_2$, as required for the bound hypervector to encode $p_1 + p_2$. \textbf{b)} The overlap between the bound hypervectors and the hypervectors $\hv{x}(p_1 + p_2)$ representing the sum, which follows (\ref{eqn:sm_lcc_bind_overlaps}).}
\label{fig:sm_lcc_bind_addition}
\end{figure}

We are searching for a relation of the form
\begin{equation}
    \hv{x}(p) \LCC \hv{x}(q) \approx \hv{x}(p + q) \LCC \hv{x}(0)
\label{eqn:sm_lcc_bind}
\end{equation}
where $\LCC$ is local circular convolution (LCC) binding operation \cite{frady_variable_2023, laiho_high-dimensional_2015}. This is important for enabling arithmetic to be performed with hypervector representations, without having to first decode into symbolic form, and thus retaining the benefits of distributed representations and possible superposition throughout. In order for this relation to hold, we require that the offsets $\theta_i$ are ordered. Using the block index $m = \lfloor i /L \rfloor$ and within-block index $n = \mathrm{mod}_L \: i$, such that $i = n + Lm$, the offsets are thus given by
\begin{equation}
    \theta_{m,n} = \theta_m - n
\end{equation}
where $\theta_m \overset{d}{=} \mathcal{U}_{[0,L]}$ is a uniformly-distributed offset specific to each block.
Then, the active index in any block is given by
\begin{equation}
    n(p) = \lfloor \mathrm{mod}_L (w_m p + \theta_m) \rfloor
\end{equation}
To probe the validity of (\ref{eqn:sm_lcc_bind}), we are interested whether, or more specifically with what probability, the following holds
\begin{equation}
    n(p) +_L n(q) = n(p+q) +_L n(0)
\end{equation}
where $+_L$ is addition modulo $L$. This is true when 
\begin{equation}
    \lfloor \omega_m p  + \theta_m \rfloor +_L \lfloor \omega_m q  + \theta_m \rfloor = \lfloor \omega_m (p+q)  + \theta_m \rfloor +_L \lfloor \theta_m \rfloor
\end{equation}
Since the integer component of each summed term doesn't affect whether the equality holds, we can simplify this equality by considering only the decimal part of each term
\begin{equation}
    \lfloor p' + \theta' \rfloor  +_L  \lfloor q' + \theta' \rfloor =  \lfloor p' + q' + \theta' \rfloor 
\label{eqn:SM_lcc_condition}
\end{equation}
where $p', q', \theta' \in [0,1)$ are the decimal part of their respective quantities, via $p' = p \omega_m - \lfloor p \omega_m \rfloor$ and $\theta' = \theta_m - \lfloor \theta_m \rfloor$. If either $p' =0 $ or $q' = 0$ in all blocks, then (\ref{eqn:SM_lcc_condition}) is true in all blocks. Thus, trivially we have 
\begin{equation}
\hv{x}(p) \LCC \hv{x}(0) = \hv{x}(p) \LCC \hv{x}(0)
\end{equation}
or alternatively
\begin{equation}
\hv{x}(p) \LCC \hv{x}(0)  \LCC^{-1} \hv{x}(0) = \hv{x}(p)
\end{equation}
where $\LCC^{-1}$ is the inverse-LCC binding operation. For vectors with one index $n$ active in each block, the hypervector where only the $L-n$'th index is active is the corresponding inverse element under LCC binding. In the general case, the inverse is computed by Fourier transform \cite{frady_computing_2022}.
If $p = q$, then $p' = q'$ in all blocks and the equality becomes
\begin{equation}
    2 \lfloor p' + \theta' \rfloor  =  \lfloor 2 p' + \theta' \rfloor
    \label{eqn:sm_lcc_p_equal_q}
\end{equation}
If $\omega_m p\gg 1$, then the decimal component $p'$ is well approximated as being sampled from the $[0,1)$ uniform distribution. Since $\theta'$ is also (independently) sampled from $\mathcal{U}_{[0,L]}$, then (\ref{eqn:sm_lcc_p_equal_q}) holds with probability $\frac{1}{2}$. For the case that $p = -q$, the equality to be satisfied is
\begin{equation}
    \lfloor p' + \theta' \rfloor + \lfloor \theta' - p' \rfloor = 0
\end{equation}
which is also true with probability $\frac{1}{2}$ when marginalising over $\theta'$ and $p'$. Hence,
\begin{equation}
\Big( \hv{x}(p) \LCC \hv{x}(\pm p) \LCC^{-1} \hv{x}(0) \Big) \cdot \hv{x}(p \pm p) \approx \frac{1}{2} \frac{N}{L} \quad \text{if} \quad \wma p \gg 1
\end{equation}
In the more general case, that $p$ is ``far away'' from $q$ ($\omega | p - q | \gg 1$ in most blocks), and neither are close to zero, then we can approximate $p'$ and $q'$ as both being drawn independently from a uniform distribution over $[0,1)$. Then, (\ref{eqn:SM_lcc_condition}) is true with probability $\frac{2}{3}$. These results are summarised by
\begin{equation}
\frac{L}{N}\Big( \hv{x}(p) \LCC \hv{x}(q) \LCC^{-1} \hv{x}(0) \Big) \cdot \hv{x}(p + q) \approx
\begin{cases}
1 & \quad \text{if} \quad  \wma p \approx 0 \text{  or  }  \wma q \approx 0 \\
\frac{1}{2} & \quad \text{else if} \quad \wma p \approx \pm  \wma q \\
\frac{2}{3} & \quad \text{otherwise}
\end{cases}
\label{eqn:sm_lcc_bind_overlaps}
\end{equation}
and demonstrated in Fig. \ref{fig:sm_lcc_bind_addition}. An additional clean-up step with the help of an associative memory would then be needed, if one wanted to recover the summed hypervector $\hv{x}(p + q)$ exactly .

\FloatBarrier

\begin{table*}
\centering
\begin{tabular}{llll}
    \toprule
    \textbf{Parameter} & \textbf{Description} & \textbf{Value}& \textbf{Notes} \\
    \midrule
    \multicolumn{4}{c}{\textbf{Model from \cite{zhang_representation_1996}}} \\
    \midrule
    $N$ & Number of neurons & 4096 & \\
    $\lambda_0$ & Weight regularisation param. & 1e-3 & As originally given \\
    $A$ & Ideal firing rate param. & \qty{1}{\Hz} & As originally given \\
    $f_\text{max}$ & Ideal firing rate param. & \qty{40}{\Hz} & As originally given \\
    $K$ &  Ideal firing rate param.  & 8 & As originally given \\
    $\beta$ & Activation func. param  & 0.8 & As originally given \\
    $a$ & Activation func. param. & 6.34 & As originally given  \\
    $b$ & Activation func. param. & 10 & As originally given \\
    $c$ & Activation func. param. & 0.5 & As originally given \\
    $\tau$ & Neuron time constant & \qty{0.25}{\ms} & Reduced from \qty{10}{\ms} originally \\
    $w_\text{scale}$ & Weight scale factor & 1.2 & Added to prevent bump from ``dying'' \\
    $\delta t$ & Simulation time step & \qty{0.05}{\ms} & \\
    \midrule
    \multicolumn{4}{c}{\textbf{Model from \cite{kilpatrick_wandering_2013}}} \\
    \midrule
    $N$ & Number of neurons & 4096 & \\
    $\sigma_\text{kp}$ & Structured heterogeneity strength & 0.2 & As originally given \\
    $\theta_\text{kp}$ & Neuron Heaviside thresh. & 0.5 & As originally given \\
    $\tau$ & Neuron time constant & \qty{0.25}{\ms} & Omitted in original work \\
    $\delta t$ & Simulation time step & \qty{0.05}{\ms} & \\
    \bottomrule
\end{tabular}
\caption{Parameters used in the comparison models from \textcite{zhang_representation_1996} and \textcite{kilpatrick_wandering_2013}.}
\label{tab:SM_parameters}
\end{table*}

\vspace{3em}

\printbibliography[title={Supplementary references}]

\newpage

\end{refsection}

\end{document}